\documentclass[10pt,journal,compsoc]{IEEEtran}

\ifCLASSOPTIONcompsoc
  \usepackage[nocompress]{cite}
\else
  \usepackage{cite}
\fi
\ifCLASSINFOpdf
\else
\fi

\pdfoutput=1

\usepackage{subfigure}
\usepackage{xcolor}
\usepackage{footnote}
\usepackage{multirow}
\usepackage{times}
\usepackage{epsfig}
\usepackage{graphicx}
\usepackage{amsmath}
\usepackage{amssymb}
\usepackage{fixltx2e}
\usepackage{color}
\usepackage{algorithm}
\usepackage{algpseudocode}
\usepackage{hyperref}
\usepackage[switch]{lineno}

\newcommand\red[1]{{\color{black}#1}}
\graphicspath{{./figures/}{./figures/draw/}{./figures/fobj_eval/}}

\def\x{{\mathbf x}}

\def\v{{\mathbf v}}
\def\h{{\mathbf h}}
\def\a{{\mathbf a}}

\def\w{{\mathbf w}}
\def\X{{\mathbf X}}
\def\C{{\mathbf C}}

\def\B{{\mathbf B}}
\def\0{{\mathbf 0}}
\def\O{{\mathcal O}}

\def\R{{\mathbb R}}

\newcommand\norm[1]{\left\lVert#1\right\rVert}

\newtheorem{theorem}{Theorem}[section]
\newtheorem{lemma}[theorem]{Lemma}

\newtheorem{corollary}[theorem]{Corollary}
\newtheorem{definition}[theorem]{Definition}

\begin{document}
%\linenumbers
\title{Embedding based on function approximation for large scale image search}
%\title{Function approximation-based embedding for large scale image search}

%\author{Michael~Shell,~\IEEEmembership{Member,~IEEE,}
%        John~Doe,~\IEEEmembership{Fellow,~OSA,}
%        and~Jane~Doe,~\IEEEmembership{Life~Fellow,~IEEE}% <-this % stops a space
%\IEEEcompsocitemizethanks{\IEEEcompsocthanksitem M. Shell was with the Department
%of Electrical and Computer Engineering, Georgia Institute of Technology, Atlanta,
%GA, 30332.\protect\\
%% note need leading \protect in front of \\ to get a newline within \thanks as
%% \\ is fragile and will error, could use \hfil\break instead.
%E-mail: see http://www.michaelshell.org/contact.html
%\IEEEcompsocthanksitem J. Doe and J. Doe are with Anonymous University.}% <-this % stops an unwanted space

%% uncomment for showing author list
\author{Thanh-Toan Do and 
        Ngai-Man Cheung
        
\IEEEcompsocitemizethanks{\IEEEcompsocthanksitem Thanh-Toan Do is with the Singapore University of Technology and Design, Singapore and the University of Adelaide, Australia.
%\IEEEcompsocitemizethanks{\IEEEcompsocthanksitem Thanh-Toan Do is with the Australian Centre for Robotic Vision, the University of Adelaide, Australia.
\protect\\
E-mail: thanh-toan.do@adelaide.edu.au
% <-this % stops an unwanted space
\IEEEcompsocthanksitem Ngai-Man Cheung is with the Singapore University of Technology and Design, Singapore.\protect\\
E-mail: ngaiman\_cheung@sutd.edu.sg
}
}

%{Shell \MakeLowercase{\textit{et al.}}: Bare Demo of IEEEtran.cls for Computer Society Journals}

\IEEEtitleabstractindextext{%

%\begin{IEEEkeywords}
%Computer Society, IEEE, IEEEtran, journal, \LaTeX, paper, template.
%\end{IEEEkeywords}
}

\maketitle

\IEEEdisplaynontitleabstractindextext

\IEEEpeerreviewmaketitle

\begin{abstract}
The objective of this paper is to design an embedding
method that maps local features describing an image (e.g. SIFT) to a higher
dimensional representation useful for the image retrieval problem.
First, motivated
by the relationship between the linear approximation of a nonlinear
function in high dimensional space and the state-of-the-art feature
representation
used in image retrieval, i.e., VLAD, we propose a new
approach for the approximation. The embedded vectors resulted by the
function approximation process are then aggregated to form a single
representation for image retrieval. Second, in order to make
the proposed embedding method applicable to large scale problem, we
further derive its fast version in which the embedded vectors can be
efficiently computed, i.e., in the closed-form. We compare the
proposed embedding
methods with the state of the art in the context of image search
under various settings: when the images
are represented by medium length vectors, short vectors, or binary
vectors. The experimental results show that the proposed embedding
methods outperform existing the state of the art on the
standard public image retrieval benchmarks.
\end{abstract}

\section{Introduction}
%man
Finding a single vector representing a set of local descriptors extracted from an image is an important problem in computer vision. This single vector representation provides several important benefits. First, it contains the power of local descriptors, such as a set of SIFT descriptors~\cite{SIFT_Lowe}. Second, the representation vectors can be 
used in image retrieval problem (comparison using standard metrics such as Euclidean distance), or in classification problem (input to robust classifiers such as SVM).
%either compared with standard distances %such as $L_2, cosin$ usually 
% used in image retrieval problem or %these presented vectors 
% used by robust classification methods such as SVM in a classification problem. 
Furthermore, they can be readily used with the recent advanced indexing techniques~\cite{herve_pami2011, kalantidis2014locally} for large scale image retrieval problem. 

There is a wide range of methods for finding a single vector to represent a set of local vectors proposed in the literature: bag-of-visual-words (BoW)~\cite{Sivic03}, Fisher vector~\cite{DBLP:conf/cvpr/PerronninD07}, vector of locally aggregated descriptor (VLAD)~\cite{herve_cvpr2010} and its improvements~\cite{DBLP:conf/mm/DelhumeauGJP13, DBLP:conf/cvpr/ArandjelovicZ13}, residual enhanced visual vector \cite{DBLP:journals/ieeemm/ChenG14},  super vector coding~\cite{DBLP:conf/eccv/ZhouYZH10}, 
vector of locally aggregated tensor (VLAT)~\cite{DBLP:conf/icip/PicardG11,DBLP:journals/ieeemm/NegrelPG13} which is higher order (tensor) version of VLAD, triangulation embedding (Temb)~\cite{DBLP:conf/cvpr/JegouZ14}, sparse coding~\cite{Olshausen97sparsecoding}, local coordinate coding (LCC)~\cite{DBLP:conf/nips/YuZG09}, locality-constrained linear coding~\cite{DBLP:conf/cvpr/WangYYLHG10} which is fast version of LCC, local coordinate coding using local tangent (TLCC)~\cite{DBLP:conf/icml/YuZ10} which is higher order version of LCC. Among these methods, VLAD~\cite{jegou11d} and VLAT~\cite{DBLP:journals/ieeemm/NegrelPG13} are well-known embedding methods used in image retrieval problem~\cite{jegou11d, DBLP:journals/ieeemm/NegrelPG13} while TLCC~\cite{DBLP:conf/icml/YuZ10} is one of the successful embedding methods used in image classification problem. 

VLAD is designed for image retrieval problem while TLCC is designed for image classification problem. They are derived from different motivations: for  VLAD, the motivation is to characterize the distribution of residual vectors over Voronoi cells learned by a quantizer; for TLCC, the motivation is to \textit{linearly approximate}\footnote{\red{The ``linear approximation" means that the nonlinear function $f(\x)$ defined on $\R^d$ is approximated by a linear function $\w^T\phi(\x)$ (w.r.t. $\phi(\x)$) defined on $\R^D$ where $D>d$.}} a nonlinear function in high dimensional space. Despite these differences, we show that VLAD is actually a simplified version of TLCC based on our original analysis.  The consequence of this analysis is significant: we can depart from the idea of linear approximation of function to develop powerful embedding methods for the image retrieval problem.

In order to compute the single representation, all aforementioned  methods include two main steps in the processing: embedding and aggregating. The embedding step uses a visual vocabulary (a set of anchor points) to map each local descriptor to a high dimensional vector while the aggregating step converts the set of mapped high dimensional vectors to a single vector. This paper focuses on the first step. In particular, we develop a new embedding method which can be seen as the generalization of TLCC and VLAT. 

%There are two main approaches proposed for second step. They are sum-pooling~\cite{DBLP:conf/cvpr/LazebnikSP06} and max-pooling~\cite{DBLP:conf/cvpr/YangYGH09}. At beginning, max-pooling can only be applied to BoW or sparse coding features. Recently, there are some new researches~\cite{DBLP:conf/cvpr/JegouZ14, DBLP:conf/cvpr/MurrayP14} making max-pooling can be applied to general features such as VLAD or Fisher vector.

In the next sections, we first present a brief description of TLCC.  Importantly, we derive the relationship between TLCC and VLAD. We then present our motivation for designing the new embedding method.

\subsection{TLCC}
TLCC~\cite{DBLP:conf/icml/YuZ10} is designed for image classification problem. Its goal is to linearly approximate a  nonlinear smooth function $f(\x)$, e.g. a nonlinear classification function, defined on a high dimensional feature space $\R^d$. 
\red{Note that $f$ is an implicit function and we do not need to know its form explicitly.}
TLCC's approach finds an embedding scheme $\phi$: $\R^d \to \R^D$ mapping each $\x \in \R^d$ as
\begin{equation}
\x \mapsto \phi(\x)
\label{eq:mapphi}
\end{equation}
such that $f(\x)$ can be well approximated by a linear function, namely $\mathbf{w}^T\phi(\x)$.
To solve above problem, TLCC's authors relied on the idea of coordinate coding defined below. They showed that with a sufficient selection of coordinate coding, the function $f(\x)$ can be linearly approximated. 

\textit{
\begin{definition}{Coordinate Coding~\cite{DBLP:conf/nips/YuZG09}}\\
\label{def:coordinate_coding}
A coordinate coding of a point $\x \in \R^d$ is a pair $(\gamma(\x), \C)$ \footnote{$\C$ is same for all $\x$.}, where $\C = [\v_1,\dots,\v_n] \in \mathbb{R}^ {d\times n}$ is a set of $n$ anchor points (bases), and $\gamma$ is a map of $\x \in \mathbb{R}^d$ to $\gamma(\x)= \left[ \gamma_{\v_1}(\x), \dots,\gamma_{\v_n}(\x) \right]^T \in \R^n$ such that 
\begin{equation}
\sum_{j=1}^{n} \gamma_{\v_j}(\x) = 1
\label{eq:cc1}
\end{equation}
It induces the following physical approximation of $\x$ in $\mathbb{R}^d$:
\begin{equation}
\x' = \sum_{j=1}^{n} \gamma_{\v_j}(\x)\v_j
\end{equation} 
A good coordinate coding should ensure that $\x'$ closes to $\x$~\footnote{Although the reconstruction error condition for a good coordinate coding, i.e, $\x'$ closes to $\x$, is not explicit mentioned in the original definition of coordinate coding, it can be inferred from objective functions of LCC~\cite{DBLP:conf/nips/YuZG09} and TLCC~\cite{DBLP:conf/icml/YuZ10}.}. %In this paper, we consider the definition with this additional condition}.
\end{definition}
}
%TLCC's approach assume that $f$ is $(\alpha, \beta, \nu)$ Lipschitz smooth. 

%It means that for all $\x, \y \in R^d$, we have
%\begin{equation}
%\left| f(\x) - f(\y) \right| \le \alpha \norm{\x-\y}_{2}
%\end{equation} 
%and,
%\begin{equation}
%\left|f(\x) - f(\y) - \nabla f(\y)^T(\x-\y)\right| \le \beta \norm{\x-\y}_{2}^2
%\end{equation}
%and,
%\begin{eqnarray}
%{} && \left| f(\x) - f(\y) - \frac{1}{2} \left( \nabla f(\x) + \nabla f(\y) \right)^T (\x - \y) \right| \nonumber \\
%{} &\le& \nu \norm{\x-\y}_{2}^3
%\end{eqnarray}

Let $(\gamma(\x), \C)$ be coordinate coding of $\x$. Under assumption that $f$ is $(\alpha, \beta, \nu)$ Lipschitz smooth, the authors showed (in lemma 2.2~\cite{DBLP:conf/icml/YuZ10}) that, for all $\x\in R^d$
\begin{eqnarray}
{} & & \left| f(\x) - \sum_{j=1}^{n} \gamma_{\v_j}(\x) \left( f(\v_j) + \frac{1}{2}\nabla f(\v_j)^T (\x - \v_j) \right) \right| \nonumber \\
{} &\le& \frac{1}{2}\alpha \norm{\x -\x'}_{2} + \nu \sum_{j=1}^{n}|\gamma_{\v_j}(\x)|\norm{\x -\v_j}_{2}^3
\label{eq:tlcc}
\end{eqnarray}

To ensure a good approximation of $f(\x)$, the authors minimize the RHS of~(\ref{eq:tlcc}). %\footnote{In practical, they minimize the function $ \norm{\x -\x'}_{2}^2 + \mu \sum_{j=1}^{n}|\gamma_{\v_j}(\x)|\norm{\x -\v_j}_{2}^3$}. 
Equation (\ref{eq:tlcc}) means that the function $f(\x)$ can be linearly approximated by $\w^T \phi(\x)$ where $\w = \left[\frac{1}{s}f(\v_j); \frac{1}{2}\nabla f(\v_j) \right]_{j=1}^n$ and TLCC embedding $\phi(\x)$ defined as
\begin{equation}
\phi(\x) = \left[s\gamma_{\v_j}(x); \gamma_{\v_j}(x)(\x - \v_j) \right]_{j=1}^{n} \in \R^{n(1+d)}
\label{eq:phi_tlcc}
\end{equation}
where $s$ is a nonnegative constant. 

\subsection{TLCC as generalization of VLAD}
\label{subsec:tlcc_vlad}
Although TLCC is designed for classification problem and its motivation is different from VLAD, TLCC can be seen as a generalization of VLAD. % being state-of-the-art features in image retrieval. 
%From ~(\ref{eq:tlcc}), to ensure a good approximation of $f(\x)$, we should select $(\gamma(\x), \C)$ such that the LHS of ~(\ref{eq:tlcc}) should be small. 
Specifically, if we add the following constraint to $\gamma(\x)$
\begin{equation}
%{}&&\gamma_{\v_j}(\x) \ge 0 \label{eq:km1}\\
\norm{\gamma(\x)}_{0} = 1 \label{eq:km2}
\end{equation}
then we have $\x \approx \x' = \v_{*}$. The RHS of~(\ref{eq:tlcc}) becomes 
\begin{equation}
\frac{1}{2}\alpha \norm{\x - \v_{*}}_{2} + \nu \norm{\x -\v_{*}}_{2}^3 
\label{eq:km3}
\end{equation}
%$\phi(\x)$ in~(\ref{eq:phi_tlcc}) becomes 
%\begin{equation}
%\phi(\x) = \left[0,\dots, 0,s,(\x - \v_{*})^T,0,\dots,0 \right]^T \in \R^{n(1+d)}
%\end{equation}
where $\v_{*}$ is anchor point corresponding to the nonzero element of $\gamma(\x)$. 
One solution for minimizing~(\ref{eq:km3}) under constraints~(\ref{eq:cc1}) and~(\ref{eq:km2}) is K-means algorithm. When K-means is used, we have
\begin{equation}
\v_{*} = \underset{\v \in \C} {\mathrm{argmin}} \norm{\x-\v}_{2}
\end{equation}
where $\C$ is set of anchor points learned by K-means. 
%We also have $\phi(\x) = [0,\dots,0,s,(\x - \v_{*}),0,\dots,0] \in \R^{n(1+d)}$. 
%If we choose $s=0$ (so we can remove this zero element) and we also remove zero elements caused by $\gamma_{\v_j}(x)$ (where $\v_j \neq \v_{*}$), $\phi(\x) = \left[0,\dots, 0,(\x - \v_{*})^T,0,\dots,0 \right]^T \in \R^{nd}$ will become VLAD. 
Now, considering~(\ref{eq:phi_tlcc}), 
%if we choose $s=0$ and we remove zero elements attached with $s$, 
if we ignore its first term, i.e., removing components attached with $s$,  we have 
$\phi(\x) = \left[0,\dots, 0,(\x - \v_{*})^T,0,\dots,0 \right]^T \in \R^{nd}$ which  becomes \red{the embedding used in VLAD.} 

\subsection{Motivation for designing new embedding method}
%Our motivation for designing new embedding $\phi(\x)$ for function approximation comes from following observation. 
The relationship between TLCC and VLAD suggests that if we can find $\phi(\x)$ such that $f(\x)$ can be well linearly-approximated ($f(\x) \approx \w^T\phi(\x)$), then $\phi(\x)$ can be a powerful feature for image retrieval problem. In TLCC's approach, by departing from assumption that $f$ is $(\alpha, \beta, \nu)$ Lipschitz smooth, $f$ is approximated using only by its first order approximation at the anchor points, i.e., $f$ is approximated as sum of weighted tangents at anchor points. 
%It is not straightforward to use the TLCC framework to have a better approximation, i.e., approximation of $f$ using its second order or higher order approximation at anchor points. 
\red{In their work \cite{DBLP:conf/icml/YuZ10}, the authors do not show how to generalize the approximation using higher order information.}

%Therefore, in this paper, we propose to use the idea of Taylor expansion for function approximation.  Using our proposed framework, it is more straightforward to achieve a higher order approximation of $f$ at the anchor points.
\red{In this paper, we propose to use the idea of Taylor expansion for function approximation. We propose an general formulation which allows to linearly approximate a nonlinear function using not only first order but also higher order information.} 

The embedded vectors, resulted by the proposed function approximation process, are used as new image representations in our image retrieval framework. In following sections, we note our \textbf{F}unction \textbf{A}pproximation-based \textbf{emb}edding method as \textbf{FAemb}. 
In order to facilitate the use of the embedded features in large scale image search problem, we further derive its fast version. The main idea is to relax the function approximation bound such that the embedded features can be efficiently computed, i.e., in an analytic form. 
The proposed embedding methods are evaluated in image search context under various settings: when the images are represented by medium length vectors, short vectors, or binary vectors. The experimental results show that the proposed methods give a performance boost over the state of the art on the standard public image retrieval benchmarks. 

%A preliminary version of this article which introduces FAemb method has appeared in 
Our previous work introduced FAemb method in
\cite{do_cvpr15}. This paper discusses substantial extension to our previous work:
We detail the computational complexity of FAemb  (Section \ref{subsec:offline-faemb}). We propose the fast version of FAemb, i.e., FAST-FAemb (Section \ref{subsec:fast-faemb}). We add a number of new experiments, i.e., results on large scale datasets  (Oxford105k and Flickr1M), results when the single representation is compressed to compact binary codes (see Section \ref{subsec:compare_soa}). We also add new experiments when Convolutional Neural Networks (CNN) features are used instead of SIFT local features to describe the images; the comparison to the recent CNN/deep learning-based image retrieval is also provided (Section \ref{subsec:cnn}). 

The remaining of this paper is organized as follows. Section~\ref{sec:pre} presents related background. Section~\ref{sec:proposed} presents FAemb embedding method. Section \ref{subsec:fast-faemb} presents the fast version of FAemb. Section~\ref{sec:exp} presents experimental results. Section~\ref{sec:concl} concludes the paper. %The implementation of the proposed embedding methods is released at \textit{\url{http://www.wikibooks.org}}.

\section{Preliminaries}
\label{sec:pre}
In this section, we review related background to prepare for detail discussion of our new embedding method in Section~\ref{sec:proposed}.\\
\\
%\subsection{Taylor's theorem for high dimensional variables}
\textbf{Taylor's theorem for high dimensional variables}
\textit{
\begin{definition}{Multi-index~\cite{folland2002advanced}: }%{Multi-index~\cite{DBLP:tibkat_590961152}:}
A multi-index is a $d$-tuple of nonnegative integers. 
Multi-indices are generally denoted by $\alpha$:
\begin{equation}
\alpha = (\alpha_1, \alpha_2, \dots, \alpha_d) \nonumber
\end{equation}
 where $ \left( \alpha_j \in \{0,1,2,...\}\right)$.
If $\alpha$ is a multi-index, we define
\begin{equation}
\vert \alpha \vert = \alpha_1 + \alpha_2 + \dots + \alpha_d; \alpha! = \alpha_1!\alpha_2!\dots\alpha_d! \nonumber 
\end{equation}
\begin{equation}
\x^\alpha = {x_1}^{\alpha_1}{x_2}^{\alpha_2}\dots{x_d}^{\alpha_d} \nonumber
\end{equation}
\begin{equation}
\partial^\alpha f(\x) = \frac{\partial^{\vert \alpha\vert} f(\x)}{\partial^{\alpha_1}(x_1) \partial^{\alpha_2}(x_2) \dots \partial^{\alpha_d}(x_d)} \nonumber
\end{equation}
where $\x = \left( x_1, x_2,\dots x_d\right)^T \in \mathbb{R}^d$
\end{definition}
}

\textit{
\begin{theorem}
\emph{(Taylor's theorem for high dimensional variables \cite{folland2002advanced})}
%Suppose $f$: $\mathbb{R}^n \to \mathbb{R}$ of class of $C^{k+1}$ on a open convex set $S$. 
%If $\a \in S$ and $\a + \h \in S$, then 
Suppose $f$: $\mathbb{R}^d \to \mathbb{R}$ of class of $C^{k+1}$ \footnote{It means that all partial derivatives of $f$ up to (and including) order $k+1$ exist and continuous.} on $\R^d$. 
If $\a \in \R^d$ and $\a + \h \in \R^d$, then
\begin{equation}
f(\a + \h) = \sum_{ | \alpha | \le k} \frac{\partial^{\alpha} f(\a)}{\alpha!}\h^\alpha + R_{\a,k}(\h) 
\end{equation}
where $R_{\a,k}(\h)$ is Lagrange remainder given by
\begin{equation}
R_{\a,k}(\h) = \sum_{ | \alpha | = k+1} \partial^{\alpha} f(\a+c\h)\frac{\h^\alpha}{\alpha!}
\end{equation}
for some $c\in (0, 1)$.
\label{taylor}
\end{theorem}
}
\textit{
\begin{corollary}
%If $f$ is of class of $C^{k+1}$ on $S$ and $\vert \partial^{\alpha} f( \x )\vert \le M$ for  $\x \in S$ and $| \alpha | = k+1$, then
If $f$ is of class of $C^{k+1}$ on $\R^d$ and $\vert \partial^{\alpha} f( \x )\vert \le M$ for  $\x \in \R^d$ and $| \alpha | = k+1$, then
\begin{equation}
\left| R_{\a,k}(\h) \right| \le \frac{M}{(k+1)!}  \norm{\h}_1^{k+1} 
\end{equation}
%where
%$\Vert h \Vert = \vert h_1 \vert + \vert h_2 \vert +\dots + \vert h_n\vert$.
\label{bound_remainder}
\end{corollary}
}
The proof of Corollary~\ref{bound_remainder} is given in~\cite{folland2002advanced}%~\cite{DBLP:tibkat_590961152}.

%\paragraph{The condition for the existing of $M$:} 
%\begin{lemma}
%If $\R^d \to \R$ is of class of $C^{k+1}$ and $\nabla^k f(\x)$ is Lipschitz continuous, then exist $M$ such that $|\partial^{\alpha}f(\x)| \le M$ for $\x\in R^d$ and $|\alpha| = k+1$
%\end{lemma}

\section{Embedding based on function approximation (FA{emb})}
\label{sec:proposed}
%In this section, we introduce our embedding method. It is inspired from function approximation based on Taylor's theorem represented in previous section. 
\subsection{Derivation of FAemb}
Our embedding method is inspired from the function approximation based on Taylor's theorem. It  comes from the following lemma.
\textit{
\begin{lemma}
If $f$: $\R^d \to \R$ is of class of $C^{k+1}$ on $\R^d$ and $\nabla^k f(\x)$ is Lipschitz continuous with constant $M>0$ and $(\gamma(\x), \C)$ is coordinate coding of $\x$, then
\begin{eqnarray}
{} & &\left| f(\x) - \sum_{j=1}^{n} \gamma_{\v_j}(\x)  \sum_{ | \alpha | \le k} \frac{\partial^{\alpha} f(\v_j)}{\alpha!} \left( \x -  \v_j \right)^\alpha \right|\nonumber \\
{} &\le & \frac{M}{(k+1)!} \sum_{j=1}^{n} |\gamma_{\v_j}(\x)| \norm{\x - \v_j}_{1}^{k+1}
\label{eq:lemma}
\end{eqnarray}
\label{lemma_1}
\end{lemma}
}
The proof of Lemma~\ref{lemma_1} is given in Appendix~\ref{proof_lemma_1}.

If $k = 1$, then~(\ref{eq:lemma}) becomes %http://math.stackexchange.com/questions/5076/what-does-%E2%88%87-upside-down-triangle-symbol-mean-in-this-problem
\begin{eqnarray}
\label{eq:taylor_k1}
{} & & \hspace{-3.5em}\left| f(\x) - \sum_{j=1}^{n} \gamma_{\v_j}(\x) \left( f(\v_j) + \nabla f(\v_j)^T \left( \x -  \v_j \right) \right) \right| \nonumber \\
{} &\le& \frac{M}{2} \sum_{j=1}^{n} |\gamma_{\v_j}(\x)| \norm{\x - \v_j}_{1}^2
\end{eqnarray}
In the case of $k=1$, $f$ is approximated as sum of its weighted tangents at anchor points. 

If $k = 2$, then~(\ref{eq:lemma}) becomes
\begin{eqnarray}%http://tex.stackexchange.com/questions/49890/linebreak-between-left-and-right
\small
\label{eq:taylor_k2}
{} & & \hspace{-3.5em}\left| f(\x) - \sum_{j=1}^{n} \gamma_{\v_j}(\x) \Bigl( f(\v_j) + \nabla f(\v_j)^T \left( \x -  \v_j \right)+ \right. \nonumber \\
{} & & \hspace{-4em}\left. \frac{1}{2} \left( V\left(\nabla^2 f(\v_j)\right) \right)^T V\left( ( \x -  \v_j ) ( \x -  \v_j )^T \right)  \Bigr)  \vphantom{\sum_{j=1}^{n}} \right| 
\le B_{FAemb}
\end{eqnarray}
\normalsize
where 
\begin{equation}
B_{FAemb} = \frac{M}{6} \sum_{j=1}^{n} |\gamma_{\v_j}(\x)| \norm{\x - \v_j}_{1}^3
\end{equation}
and $V(\mathbf{A})$ is vectorization function flattening the matrix $\mathbf{A}$ to a vector by putting its consecutive columns into a column vector. $\nabla^2$ is Hessian matrix.

In the case of $k=2$, $f$ is approximated as sum of its weighted quadratic approximations at anchor points. \red{Note that in both (\ref{eq:taylor_k1}) and (\ref{eq:taylor_k2}), we do not need to know the explicit form of function $f$. In the rest, we put the interest on the function approximation bound}. 

In order to achieve a good approximation, the coding $(\gamma(\x), \C)$ should be selected such that the RHS of ~(\ref{eq:taylor_k1}) and ~(\ref{eq:taylor_k2}) are small enough. 
  
The result derived from~(\ref{eq:taylor_k1}) is that, with respect to the coding $(\gamma(\x), \C)$, a high dimensional nonlinear function $f(\x)$ in $\R^d$ can be approximated by linear form $\w^T\phi(\x)$ where $\w$ can be defined as $\w = \left[ \frac{1}{s}f(\v_j); \nabla f(\v_j) \right]_{j=1}^n$ %using its local tangents (at anchor points) having linear form
%\begin{equation}
%\w^T\phi(\x) =  \sum_{j=1}^{n} \gamma_{\v_j}(\x) \left( f(\v_j) + \nabla f(\v_j)^T \left( \x -  \v_j \right) \right) \nonumber
%\end{equation} 
%~(\ref{eq:taylor_k1}) further suggests that 
and the embedded vector $\phi(\x)$ can be defined as  
\begin{equation}
\phi(\x) = \left[ s\gamma_{\v_j}(\x); \gamma_{\v_j}(\x) (\x - \v_j)\right]_{j=1}^{n} \in \R^{n(1+d)}
\label{eq:phix_1}
\end{equation}
where $s$ is a nonnegative scaling factor to balance two types of codes. 

%We can see that $\phi(\x)$ in~(\ref{eq:phix_1}) has similar form as TLCC~(\ref{eq:phi_tlcc}). However~(\ref{eq:phix_1}) is resulted from function bound~(\ref{eq:taylor_k1}) which is different from TLCC's function bound~(\ref{eq:tlcc}). Therefore, the coordinate coding $(\gamma(\x),\C)$ will be different.  

In order to make a good approximation of $f$, in following sections, we put our interest on case where $f$ is approximated by using up to second-order derivatives defined by~(\ref{eq:taylor_k2}). The result derived from~(\ref{eq:taylor_k2}) is that the nonlinear function $f(\x)$ can be approximated by linear form $\w^T\phi(\x)$ where $\w$ can be defined as $\w = \left[\frac{1}{s_1} f(\v_j); \frac{1}{s_2} \nabla f(\v_j); \frac{1}{2} \left( V\left(\nabla^2 f(\v_j)\right) \right) \right]_{j=1}^n$ 
and the embedded vector $\phi(\x)$-FAemb can be defined as 
\begin{eqnarray} 
\phi(\x) = \Big[ s_1\gamma_{\v_j}(\x); s_2\gamma_{\v_j}(\x) (\x - \v_j);\hspace{5.5em} &&{} \nonumber\\
\gamma_{\v_j}(\x) V\left( ( \x -  \v_j ) ( \x -  \v_j )^T \right) \Big]_{j=1}^{n}\in \R^{n(1+d+d^2)}\hspace{0em} &&{}
\label{eq:phix_2}
\end{eqnarray}
where $s_1, s_2$ are nonnegative scaling factors to balance three types of codes.

As mentioned in the previous section, to get a good approximation of $f$, the RHS of~(\ref{eq:taylor_k2}) should be small enough. %\footnote{Because~$\frac{M}{6}$ is constant, it can be ignored in the optimization process.}. 
Furthermore, from the definition of coordinate coding~\ref{def:coordinate_coding}, $(\gamma(\x), \C)$ should ensure that the reconstruction error $\norm{\x' - \x}_{2}$ should be small. Putting these two criteria together, we find $(\gamma(\x), \C)$ which minimize the following constrained objective function 
\begin{equation}
Q(\gamma(\x),\C) =  \frac{1}{2} \norm{\x - \C\gamma(\x)}_{2}^2 + \frac{\mu}{2} \sum_{j=1}^{n} |\gamma_{\v_j}(\x)| \norm{\x - \v_j}_{1}^3  \nonumber 
\end{equation} \vspace{-0.4cm}
\begin{equation}
st.\  \mathbf{1}^T\gamma (\x) = 1
\label{eq:objective_x}
\end{equation}
where $\mu$ is the parameter that regulates to the importance of the function approximation bound in the objective function. 

Equivalently, given a set of training samples (descriptors) $\X = [ \x_1,\dots,\x_m ] \in \R^{ d\times m}$, let $\gamma_{i_j}$ be coefficient corresponding to base $\v_j$ of sample $\x_i$; $\gamma_i = [\gamma_{i_1},\dots, \gamma_{i_n}]^T \in \R^n$ be coefficient vector of sample $\x_i$; $\Gamma = [ \gamma_1,\dots,\gamma_m ] \in \R^{n\times m}$. We find $(\Gamma, \C)$ which minimize the following constrained objective function 
%\small
\footnotesize
\begin{equation}
Q_{FAemb}(\Gamma, \C) = \frac{1}{m}\sum_{i = 1}^{m} \left[\frac{1}{2} \norm{\x_i - \C\gamma_i}_{2}^2 + \frac{\mu}{2} \sum_{j=1}^{n} |\gamma_{i_j}| \norm{\x_i - \v_j}_{1}^3 \right] \nonumber 
\end{equation} \vspace{-0.1cm}
\begin{equation}
st.\  \mathbf{1}^T\gamma_i = 1, i = 1,\dots, m  
\label{eq:objective}
\end{equation}
\normalsize
\subsection{The offline learning of coordinate coding and the online embedding}
\label{subsec:offline-faemb}
\subsubsection{The offline learning of coordinate coding via alternating optimization}
In order to minimize~(\ref{eq:objective}), we propose to iteratively optimize it by alternatingly optimizing with respect to $\C$  and $\Gamma$ while holding the other fixed. 

For learning the anchor points $\C$, the optimization  is an unconstrained regularized least squares problem. We use trust-region method~\cite{coleman1996interior} to solve this problem.

For learning the coefficients $\Gamma$, the optimization is equivalent to a regularized least squares problem with linear constraint. This problem can be solved by optimizing over each sample $\x_i$ individually. To find $\gamma_i$ of each sample $\x_i$, 
we use Newton's method~\cite{boyd2004convex}. %The gradient and Hessian of objective function w.r.t. $\gamma_i$ is given in Appendix~\ref{app:derivative}.

 %~\footnote{Because the objective function involves $L_1$ norm, some methods designed for $L_1$ regularization, i.e, feature-sign search algorithm~\cite{lee2006efficient}, can be used. However, we find that the Newton's method (for computing $\Gamma$) and the trust-region method (for computing $\C$) work well in practice.}. %http://en.wikipedia.org/wiki/Newton%27s_method_in_optimization
%The gradient and Hessian of objective function w.r.t. $\C$ is given in Appendix~\ref{app:derivative}.

\begin{figure}[!t]
\centering
\includegraphics[scale=0.5]{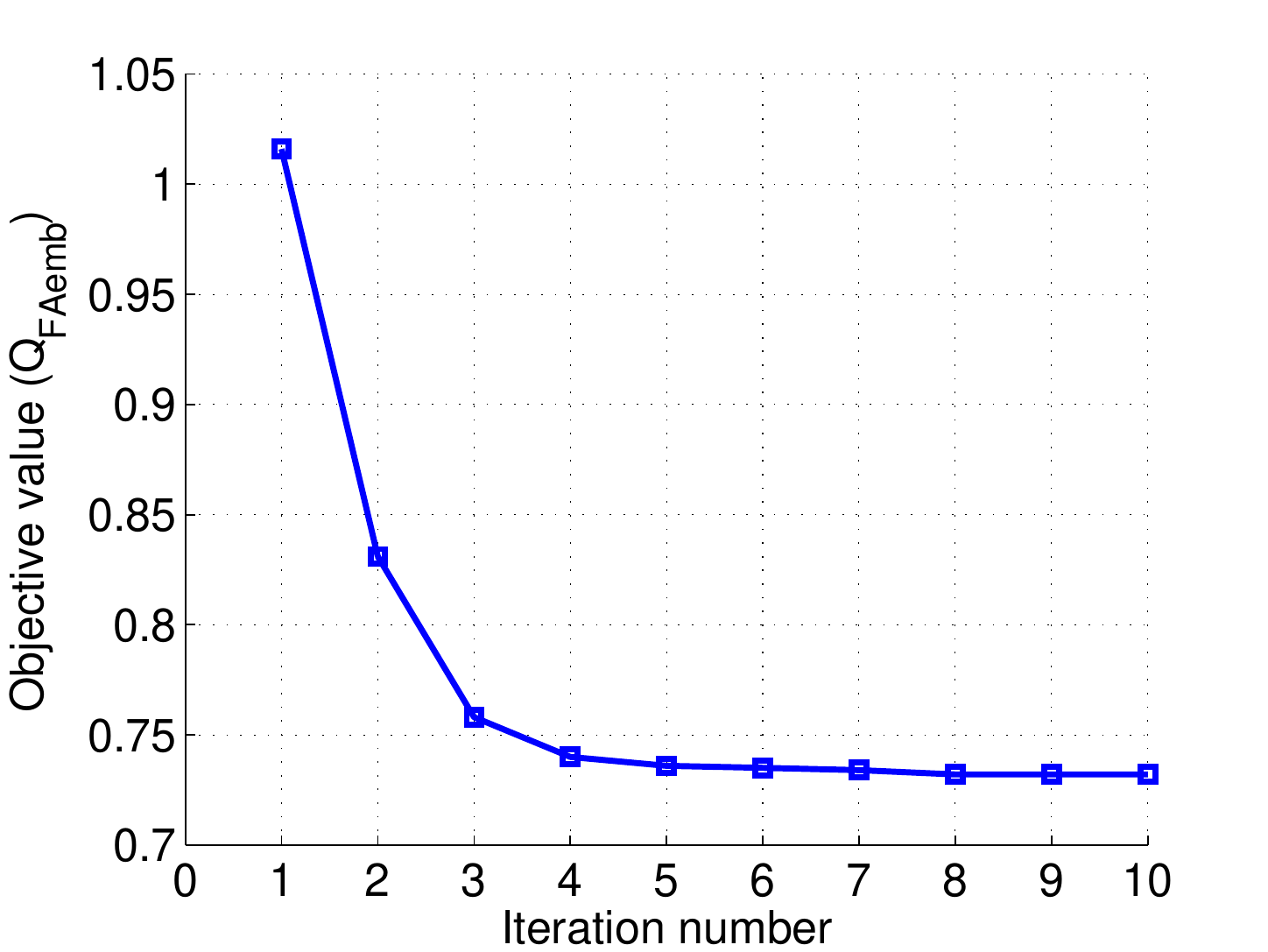}
\caption{The convergence of the Algorithm \ref{alg1}. Number of anchor points $n=8$; $\mu=10^{-2}$. 1M descriptors are used for training.}
\label{objective_value}
\end{figure}

The offline learning of coordinate coding for FAemb is summarized in the Algorithm~\ref{alg1}. In the Algorithm~\ref{alg1}, $\C^{(t)}$, $\Gamma^{(t)}$, $Q_{\textrm{FAemb}}^{(t)}$ are values of $\C$, $\Gamma$, $Q_{\textrm{FAemb}}$ at iteration $t$, respectively.
The objective function value $Q_{\textrm{FAemb}}$ after each iteration $t$ in the Algorithm~\ref{alg1} always does not increase (by the decreasing or unchanging of the objective value on both $\Gamma$ and $\C$ steps). It can also be validated that the objective function value is lower-bounded, i.e., not smaller than $0$. Those two points indicate the convergence of our algorithm. The empirical results show that the Algorithm \ref{alg1} takes a few iterations to converge. Fig. \ref{objective_value} shows the example of the convergence of the algorithm. 
\begin{algorithm}[!t]
	\footnotesize
	\caption{Offline learning of coordinate coding}
	\begin{algorithmic}[1] 
		\Require 
			\Statex $\X = \{\x_i\}_{i=1}^{m} \in \R^{d\times m}$: training data; $T$: max iteration number; $\epsilon$: convergence error.
		\Ensure 
			\Statex  $\C$: anchor points; $\Gamma$: coefficient vectors of $\X$.
			\Statex 
			\State Initialize $\C^{(0)}$ using K-means
			\For{$t = 1 \to T$}
				\State $\Gamma^{(t)} = \varnothing$
				\For{$i = 1 \to m$}
				\State Fix $\C^{(t-1)}$, compute $\gamma_i$ for $\x_i$ using Newton's method~\cite{boyd2004convex}
				\State $\Gamma^{(t)} \gets  [\Gamma^{(t)}, \gamma_i]$
				\EndFor
				\State Fix $\Gamma^{(t)}$, compute $\C^{(t)}$ using trust-region method~\cite{coleman1996interior}.
				\If{$t>1$ and $|Q_{\textrm{FAemb}}^{(t)}-Q_{\textrm{FAemb}}^{(t-1)}|<\epsilon$}
					\State break;
				\EndIf
			\EndFor
			\State Return $\C^{(t)}$ and $\Gamma^{(t)}$
    \end{algorithmic}
    \label{alg1}
\end{algorithm}

\subsubsection{The online embedding and its complexity}
\label{subsubsec:faemb_complex}
After learning anchor points $\C$, given a new descriptor $\x$, we achieve $\gamma(\x)$ by minimizing~(\ref{eq:objective_x}) using learned $\C$. From $\gamma(\x)$, we get the embedded vector  $\phi(\x)$-FAemb by using~(\ref{eq:phix_2}).\\
\\
\textbf{The computational complexity of the online embedding}\\
The computational complexity of the online embedding depends on the computation of the coefficient $\gamma(\x)$ using Newton's method and the computation of $\phi(\x)$ (given $\gamma(\x)$) using~(\ref{eq:phix_2}). It is worth noting that in our experiments, the number of anchor points $n (= 8,16,32)$ is less than the dimension $d (= 45)$ of descriptor. 
\\
\\
\textbf{Computing $\gamma(\x)$:} 
FAemb uses Newton's method~\cite{boyd2004convex} for finding $\gamma_i$. 
%The stopping criterion of Newton's method can be is maximum number iteration $m$ or is a given tolerance on the objective function.  %Let $m$ be number of iteration of Newton's method. $m$ can be defined by user or can be computed based on given tolerance $\epsilon$. 
The main cost in the $(t+1)^{th}$ iteration of Newton's method lies in (i) computing the Hessian of objective function~(\ref{eq:objective}) and (ii) computing the Newton step $\triangle \gamma_i$. 

The complexity for computing Hessian $\nabla^2 Q_{FAemb}$ of~(\ref{eq:objective}) w.r.t. $\gamma_i$ is $\mathcal{O}(n^2d)$. For finding the updating step $\triangle \gamma_i$, Newton's method solves the following equation
\begin{equation}
\small
\left[ 
\begin{array}{cc}
\nabla^2 Q_{FAemb}(\gamma^{(t)}_i)  &\mathbf{1} \\
\mathbf{1}^T				  &0
\end{array} 
\right]
\left[ 
\begin{array}{c}
\triangle \gamma_i \\
w
\end{array}
\right]
= \left[ 
\begin{array}{c}
-\nabla  Q_{FAemb}(\gamma^{(t)}_i)\\
0
\end{array}
\right]
\label{eq:newton}
\end{equation}
where $\gamma^{(t)}_i$ is solution at the $t^{th}$ iteration. 

The size of 1st, 2nd, 3rd matrices in~(\ref{eq:newton}) is $(n+1)\times (n+1)$, $(n+1)\times 1$ and $(n+1)\times 1$, respectively. So, the complexity for solving~(\ref{eq:newton}) is $\mathcal{O}(n^3)$. %~\cite{trefethen1997numerical}. 
%The Newton decrement $\lambda(\gamma_i)$ is computed by
%\begin{equation}
%\lambda(\gamma_i) = ( (\triangle \gamma_i)^T \nabla^2 Q_{FAemb}(\gamma_i)\triangle \gamma_i)^{\frac{1}{2}}
%\end{equation}
%which taking $\mathcal{O}(n^2)$ in complexity. So, the complexity for computing $\triangle \gamma_i$ and $\lambda(\gamma_i)$ is $\mathcal{O}(n^3)$. 
Overall, the complexity in one iteration of Newton's method is $\mathcal{O}(n^2d)$. 

For the stopping of Newton's method, follow~\cite{boyd2004convex}, we define a tolerance $\epsilon$ on the objective function, i.e., the algorithm is terminated when 
$Q_{FAemb}(\gamma^{(t+1)}_i, \C) - p^* \le \epsilon$, where $p^*$ denotes the optimum value of objective function; $\gamma^{(t+1)}_i$ is solution at the $(t+1)^{th}$ iteration.~\cite{boyd2004convex} showed that this stopping criterion is equivalent to $(\delta^{(t+1)})^2/2 \le \epsilon$, %trang 487
 where $\delta^{(t+1)}$ is Newton decrement at the $(t+1)^{th}$ iteration and defined by the following equation which takes only $\mathcal{O}(n)$ in complexity.
\begin{equation}
\delta^{(t+1)} =  \left( -(\nabla Q_{FAemb}(\gamma^{(t)}_i))^T \triangle \gamma_i \right)^\frac{1}{2}
\end{equation}
%Follow~\cite{boyd2004convex}, in each iteration of Newton's method, we compute square of Newton decrement $\delta(\gamma_i)^2$ defined by
%\begin{equation}
%\delta(\gamma_i)^2 = -(\triangle \gamma_i)^T \triangle \gamma_i
%\end{equation}
%The complexity in computing of $\delta(\gamma_i)^2$ is only $\mathcal{O}(n)$.
%Given a tolerance $\epsilon$ on the objective function$\epsilon = Q_{FAemb}(\gamma^{j}_i, \C) − p*$ where $p*$ denotes the optimum value of objective function.  
%The algorithm is terminated when $\delta(\gamma_i)^2/2 \le \epsilon$ 
Given a coordinate coding with 8 anchor points, a tolerance $\epsilon = 10^{-6}$, we experiment on 100k descriptors and have the observation that $k$ $\approx$ 50 iterations on average for meeting the stopping criterion\footnote{The step-size $\alpha$ for updating $\gamma_i$, i.e., $\gamma^{(t+1)}_{i} = \gamma^{(t)}_i + \alpha\triangle \gamma_i$ at each iteration is selected by empirical experiments and equals to 0.1 in our experiments.}. %It is worth noting that $k$ is bigger than both $n$ and $d$, so the complexity of FAemb for finding $\gamma_i$ is $\mathcal{O}(kn^2d)$. This is much higher than the complexity of FAST-FAemb and FAST-FAemb++. 
Overall, the complexity of FAemb for finding $\gamma_i$ is $\mathcal{O}(kn^2d)$. 
\\\\
\textbf{Computing $\phi(\x)$ (using~(\ref{eq:phix_2}), given $\gamma(\x)$):} %As $s_1$ and $s_2$ equal to 0, 
From (\ref{eq:phix_2}), the complexity mainly depends on the computing the tensor between $\x$ and $\v_j$ which takes $\mathcal{O}(d^2)$. So, the computational complexity for computing $\phi(\x)$ is $\mathcal{O}(nd^2)$. 

From above analysis, we find that the computational complexity of the whole embedding process of FAemb is dominated by the computing of $\gamma(\x)$.  

\subsection{Relationship to other methods}
The most related embedding methods to FAemb are TLCC~\cite{DBLP:conf/icml/YuZ10} and VLAT~\cite{DBLP:conf/icip/PicardG11}. 

Compare to TLCC~\cite{DBLP:conf/icml/YuZ10}, our assumption on $f$ in lemma~\ref{lemma_1} is  different from the assumption of TLCC (lemma $2.2$~\cite{DBLP:conf/icml/YuZ10}), i.e., our assumption only needs that $\nabla^k f(\x)$ is Lipschitz continuous while TLCC assumes that all $\nabla^j f(\x)$ are Lipschitz continuous, $j = 1,\dots,k$. Our objective function~(\ref{eq:objective_x}) is different from TLCC~(\ref{eq:tlcc}), i.e., we rely on $l_1$ norm of ($\x-\v_j$) in the second term  while TLCC uses $l_2$ norm; %\footnote{Note that $\x_i \approx \C\gamma_i$ and there is only a difference of constant by replacing the $L_2$ norm with its square. Hence, we can use $\norm{\x_i - \C\gamma_i}_{2}$ or $\norm{\x_i - \C\gamma_i}_{2}^2$. Both of them have same meaning.} 
we solve the constraint on the coefficient $\gamma$ in our optimization process while TLCC does not. 
 %Despite that similarity, we theoretically showed that by minimizing the constrained objective function (\ref{eq:objective_x}), $f$ can be approximated by using up to its second order derivatives. This is different from TLCC's approach because TLCC's approach approximates $f$ only using its first order derivatives. 
FAemb approximates $f$ using up to its second order derivatives while TLCC approximates $f$ only using its first order derivatives. 
 
%If we want to have features as TLCC, by combination the minimization of the LHS of (\ref{eq:taylor_k1}) and reconstruction error of coordinate coding definition, we will need to solve the following constrained objective function
%\begin{eqnarray}
%\label{eq:objective1}
%Q(\Gamma, \C) &=& \sum_{i = 1}^{m} \left[ \norm{\x_i - \C\gamma_i}_{2}^2 + \mu \sum_{j=1}^{n} |\gamma_{i_j}| \norm{\x_i - \v_j}_{2}^2 \right] \nonumber \\
%{} & & st.  \mathbf{1}^T\gamma_i = 1, i = 1,\dots, m  
%\end{eqnarray}
%Above objective function differs from TLCC by adding of the constraint and the second term of LHS is only up to second power instead of third power as TLCC. 

FAemb can also be seen as the generalization of VLAT~\cite{DBLP:conf/icip/PicardG11}.
Similar to the relationship of TLCC and VLAD presented in Section~\ref{subsec:tlcc_vlad}, if we add %constraints~(\ref{eq:km1}) and~(\ref{eq:km2})
the constraint~(\ref{eq:km2}) to $\gamma(\x)$, the objective function~(\ref{eq:objective}) will become 
\begin{eqnarray}
Q_1(\Gamma, \C) &=& \frac{1}{m}\sum_{i = 1}^{m} \left[ \frac{1}{2}\norm{\x_i - \v_{*}}_{2}^2 + \frac{\mu}{2}  \norm{\x_i - \v_{*}}_{1}^3 \right] \nonumber \\
{} & & st.\  \mathbf{1}^T\gamma_i = 1, i = 1,\dots, m  \nonumber \\
{} & & \hspace{1em}  \norm{\gamma_i}_{0} = 1, i = 1,\dots, m  
\label{eq:objective_vlat}
\end{eqnarray}
where $\v_{*}$ is anchor point corresponding to the nonzero element of $\gamma_i$. 

If we relax $l_1$ norm in the second term of $Q_1(\Gamma, \C)$ into $l_2$ norm, we can use K-means algorithm for minimizing~(\ref{eq:objective_vlat}). 
After learning $\C$ by using K-means, given an input descriptor $\x$, we have
\begin{equation}
\x \approx \v_{*} = \underset{\v \in \C} {\mathrm{argmin}} \norm{\x-\v}_{2}
\end{equation}

%Now, consider~(\ref{eq:phix_2}), if we choose $s_1=0, s_2 = 0$ and we remove zero elements attached with them, $\phi(\x) = [0, \dots, 0, \left( V\left( ( \x -  \v_{*} ) ( \x -  \v_{*} )^T \right)\right)^T,0 ,\dots,0]^T \in \R^{nd^2}$ will become VLAT. 
\red{Now, consider~(\ref{eq:phix_2}), if we ignore the first and the second terms, i.e., removing components attached with $s_1$ and $s_2$, we have: $\phi(\x) = [0, \dots, 0, \left( V\left( ( \x -  \v_{*} ) ( \x -  \v_{*} )^T \right)\right)^T,0 ,\dots,0]^T \in \R^{nd^2}$ which becomes the embedding used in VLAT.}

%In practice, to make a fair comparison between FAemb and VLAT, we assign $s_1, s_2$ in~(\ref{eq:phix_2}) to $0$. 
\red{In practice, to make a fair comparison between FAemb and VLAT, we remove the first and the second terms of ~(\ref{eq:phix_2}).} 
This makes the embedded vectors produced by two methods have same dimension. 
It is worth noting that in~(\ref{eq:phix_2}), as matrix $(\x - \v_j)(\x - \v_j)^T$ is symmetric, only the diagonal and upper part are kept while flattening it into vector. The size of VLAT and FAemb is then $\frac{nd(d+1)}{2}$.
\section{Fast embedding based on function approximation (F-FAemb)}
\label{subsec:fast-faemb}
FAemb needs an iterative optimization at the online embedding step.
While FAemb is applicable for small/medium-size datasets, it may not be suitable for large scale datasets.
% also in applications requiring fast retrieval. 
In this section, we develop the fast version of FAemb. The main idea is to find reasonable relaxation for the function approximation bound of FAemb  (i.e., the RHS of (\ref{eq:taylor_k2})) such that the coefficient vector $\gamma(\x)$ can be efficiently computed, i.e., it can be computed in a closed-form. 
\subsection{Derivation of F-FAemb}
The relaxed bound is based on the following observation
\begin{equation}
1 = \sum_{j=1}^n {\gamma_{\v_j}(\x)} \le \norm{\gamma(\x)}_1 \le n\norm{\gamma(\x)}_2^2 %\nonumber 
\end{equation}
Thus, 
\begin{equation}
\label{new_bound}
 B_{FAemb} \le \frac{nM}{6} \norm{\gamma(\x)}_2^2 \sum_{j=1}^{n}  \norm{\x - \v_j}_{1}^3
\end{equation}
We define the relaxed bound $B_{F-FAemb}$ as
\begin{equation}
\label{eq:fast_taylor_k2}
B_{F-FAemb} = \frac{nM}{6} \norm{\gamma(\x)}_2^2 \sum_{j=1}^{n}  \norm{\x - \v_j}_{1}^3
\end{equation}
~(\ref{new_bound}) means that the relaxed bound $B_{F-FAemb}$ is still upper bound of the function approximation, i.e, the LHS of~(\ref{eq:taylor_k2}). This relaxed bound allows analytic solution for the embedding as shown in Section \ref{subsec:ffaemb-learing}.

Similar to FAemb, in order to ensure a good reconstruction error (which is necessary condition for a good coordinate coding) and a good function approximation, we jointly minimize over the reconstruction error and the bound  $B_{F-FAemb}$. Specifically, the coordinate coding is learned by minimizing the following constrained objective function
\begin{eqnarray}
Q_{F-FAemb}(\Gamma, \C) = \hspace{13em} \nonumber \\
\frac{1}{m}\sum_{i = 1}^{m} \left[ \frac{1}{2}\norm{\x_i - \C\gamma_i}_{2}^2 + \frac{\mu}{2} \norm{\gamma_i}_2^2 \sum_{j=1}^{n}  \norm{\x_i - \v_j}_{1}^3 \right] \nonumber 
\end{eqnarray} \vspace{-0.3cm}
\begin{equation}
st.\  \mathbf{1}^T\gamma_i = 1, i = 1,\dots, m  
\label{eq:objective-ffaemb}
\end{equation}

\subsection{The offline learning of coordinate coding and the online embedding}
\label{subsec:ffaemb-learing}
\subsubsection{The offline learning of coordinate coding via alternating optimization}
Similar to FAemb, in order to optimize (\ref{eq:objective-ffaemb}), we  alternatingly optimize with respect to $\C$ and $\Gamma$ while holding the other fixed. 

For learning the anchor points $\C$, the optimization problem is unconstrained regularized least squares. We use trust-region method \cite{coleman1996interior} for solving. 

For computing the coefficients $\Gamma$, we can solve over each sample $\x_i$ individually. The optimization problem is equivalent to a $l_2$ regularized least squares problem with linear constraint. Thus, we achieve the closed-form for the solution.

Let $\a = \left[\norm{\x_i - \v_1}_1^3,\dots,\norm{\x_i - \v_n}_1^3\right]^T$; $a = \mathbf{1}^T\a$; $\B = (\C^T\C + a\mu\mathbf{I})^{-1}$. 
Let
\begin{equation}
\lambda = \frac{\mathbf{1}^T \B \C^T \x_i - 1}{\mathbf{1}^T \B \mathbf{1}} 
\end{equation}
We have the closed-form for the coefficient vector as 
\begin{equation}
\gamma^{F-FAemb}_i = \B (\C^T\x_i - \lambda \mathbf{1})
\label{eq:closed-fast-faemb}
\end{equation}
where $\mathbf{I}$ is identity matrix having size of $n \times n$; $\mathbf{1}$ is column vector having $n$ elements equaling to $1$.

It is worth noting that although the function bound of F-FAemb is the relaxed version of the function bound of FAemb, F-FAemb provides an optimum solution on the coefficient vector while FAemb does not. This explains for the results that the performance F-FAemb is  competitive to FAemb in our experiments. 

The algorithm for offline learning of coordinate coding of F-FAemb is similar to the one of FAemb presented in the Algorithm \ref{alg1}, excepting that (i) at line 5 of the Algorithm \ref{alg1}, we have the closed-form solution for $\gamma^{\textrm{F-FAemb}}_i$, i.e. using (\ref{eq:closed-fast-faemb}),  and (ii) the condition at line 9 is on $Q_{\textrm{F-FAemb}}$.

\subsubsection{The online embedding and its complexity}
After learning anchor points $\C$, given a new input $\x$, we use (\ref{eq:closed-fast-faemb}) for computing the coefficient vector $\gamma^{\textrm{F-FAemb}}$.  After getting the coefficient vector, the embedded vector $\phi(\x)$ is achieved by using~(\ref{eq:phix_2}). Similar to FAemb, the values of $s_1$, $s_2$ in (\ref{eq:phix_2}) are assigned to 0.
\\
\\
\textbf{Computing $\gamma(\x)$:} From (\ref{eq:closed-fast-faemb}), the computational complexity for computing $\B$, $\lambda$ and $\gamma^{\textrm{F-FAemb}}_i$ is $\mathcal{O}(n^2d)$, $\mathcal{O}(nd)$ and $\mathcal{O}(nd)$, respectively. So the overall computational complexity for computing $\gamma(\x)$ is $\mathcal{O}(n^2d)$.
\\
\\
\textbf{Computing $\phi(\x)$:} As presented in Section \ref{subsubsec:faemb_complex}, the computational complexity for computing $\phi(\x)$ is $\mathcal{O}(nd^2)$.

As in our experiments, the number of anchor points $n (= 8,16,32)$ is less than the dimension $d (= 45)$ of descriptor, the complexity of the whole embedding process of F-FAemb is dominated by the computing of $\phi(\x)$.  

%\subsection{The convergency in offline learning coordinate coding of FAemb/F-FAemb}
\subsection{The computational complexity comparison between FAemb/F-FAemb and other methods}
\begin{table}[!t] 
\centering
\caption{\footnotesize Computational complexity ($5^{th}$ row) and CPU timing in milliseconds ($6^{th}$ row) to embed a local descriptor. $d$ and $n$ are dimension of local descriptor and number of anchor points, respectively. $D$ is dimension of the embedded vector $\phi(\x)$ before truncating. Results are reported on the average of 100k descriptors.}
\begin{tabular}{|c|c|c|c|c|}
\hline
    &FAemb     &F-FAemb      &VLAT       &Fisher             \\ \hline
$d$  &45       &45          &45         &64                   \\ \hline
$n$  &16       &16          &16         &128                  \\ \hline
$D$  &$16,560$ &$16,560$    &$16,560$   &$16,384$       \\ \hline
     &$\O(n^2dk)$  &$\O(nd^2)$  &$\O(d^2)$ &$\O(nd)$        \\ \hline
     &8.32     &0.78        &0.25       &1.42               \\ \hline
\end{tabular}
\label{tab:comp-timing}
\end{table}

In this section, we compare the computational complexity to embed a local descriptor of FAemb/F-FAemb and other methods which also use high order (i.e., second order) information for embedding such as VLAT \cite{DBLP:conf/icip/PicardG11,DBLP:journals/ieeemm/NegrelPG13}, Fisher \cite{DBLP:conf/cvpr/PerronninLSP10,jegou11d}. 

The fifth row of Table \ref{tab:comp-timing} presents the asymptotic complexity (hence the constant of the complexity is different for each method). Note that, although the dimension of local descriptors ($d$) and the number of anchor points ($n$) of methods are different, the dimension of the embedded vector produced by methods are comparable. 
It is worth noting that for Fisher\cite{DBLP:conf/cvpr/PerronninLSP10,jegou11d}, although the complexity is $\O(nd)$, it has a large constant, i.e., by computing posterior probabilities, the gradient with respect to both the mean and the standard deviation.

The sixth row of Table \ref{tab:comp-timing}  presents the timing to embed a local descriptor. For Fisher, we use the implementation provided by VLFeat \cite{VLfeat}, where the  implementation is optimized with mex files. 
For standard VLAT \cite{DBLP:conf/icip/PicardG11}, we re-implement it as there is no Matlab implementation available. The experiments are run on a processor core (2.60 GHz Intel CPU). We report the CPU times which is larger than elapsed ones because CPU time accumulates all active threads. \red{It is worth noting that we measure the timing when computing the VLAT/Fisher for each local descriptor separately. The reason is presented in Section \ref{subsec:compare_4methods}}.

From $6^{th}$ row of Table \ref{tab:comp-timing}, F-FAemb is $\sim 11$ times faster than FAemb. F-FAemb is slower than VLAT while it is faster than Fisher. %, in spite of the optimized implementation of Fisher. 
 In practical, F-FAemb takes less than 1$s$ to embed an image having 1,000 local descriptors. This efficient computation allows F-FAemb to be used in large scale problems
and in applications requiring fast retrieval. It is worth noting that the embedding can be further speeded up by optimizing the implementation, i.e., using mex files. In our experiments, when using mex file implementation for computing $\phi(\x)$ (\ref{eq:phix_2}), F-FAemb takes only $0.20$ ms to embed a local descriptor.

\section{Experiments}
\label{sec:exp}
This section presents results of the proposed FAemb and F-FAemb embedding methods and compare them to the state of the art. Specifically, 
\red{in Section \ref{subsec:compare_4methods}, we compare FAemb, F-FAemb to other methods: VLAD~\cite{jegou11d}, Fisher \cite{jegou11d,DBLP:conf/cvpr/PerronninD07}, Temb~\cite{DBLP:conf/cvpr/JegouZ14} and VLAT~\cite{DBLP:conf/icip/PicardG11}
}
 when the same test bed are used. In Section \ref{subsec:compare_soa}, we compare FAemb, F-FAemb to the state of the art under various setting, i.e., when the images are represented by mid-size vectors, short vectors, or binary vectors. Furthermore, in Section \ref{subsec:cnn}, we present the results when Convolutional Neural Network features %, which is state-of-the-art image representation, 
are used as the local features to describe the images. The comparison to the recent CNN/deep learning-based image retrieval are also provided.

\subsection{Dataset and evaluation metric}
\label{subsec:dataset}
\textbf{INRIA holidays~\cite{herve_ijcv2010}} consists of 1,491 images of different locations and objects, 500 of them being used as queries. The search quality is measured by mean average precision (mAP), with the query removed from the ranked list. 
In order to evaluate the search quality on large scale, we merge Holidays dataset with 1M negative images downloaded from Flickr~\cite{herve_eccv2008}, forming the~\textbf{Holidays+Flickr1M} dataset. For this large scale dataset, following common practice~\cite{DBLP:conf/cvpr/JegouZ14}, we evaluate search quality on the short representations of the aggregated vectors. %As standardly done in the literature, for all learning stages, we use 10k images from the independent dataset Flickr60k provided with Holidays. 
For all learning stages, we use a subset from the independent dataset Flickr60k provided with Holidays. 
\\
\\
\textbf{Oxford buildings~\cite{Philbin07-cvpr-2007}} consists of 5,063 images of buildings and 55 query images corresponding to 11 distinct buildings in Oxford. Each query image contains a bounding box indicating the region of interest. When local SIFT features are used, we follow the standard protocol \cite{DBLP:conf/mm/DelhumeauGJP13,DBLP:conf/cvpr/ArandjelovicZ13, DBLP:conf/cvpr/JegouZ14}: the bounding boxes are cropped and then used as the queries. 
This dataset is often referred to as~\textbf{Oxford5k}. The search quality is measured by mAP computed over the 55 queries. Images are annotated as either relevant, not relevant, or junk, which indicates that it is unclear whether a user would consider the image as relevant or not. Following the recommended configuration~\cite{DBLP:conf/mm/DelhumeauGJP13, DBLP:conf/cvpr/JegouZ14, herve_cvpr2010}, the junk images are removed from the ranking before computing the mAP. 
In order to evaluate the search quality on large scale, Oxford5k is extended with 100k negative images~\cite{Philbin07-cvpr-2007}, forming the~\textbf{Oxford105k} dataset. For all learning stages, we use the Paris6k dataset~\cite{Philbin08_cvpr}.

\subsection{Implementation details}
\subsubsection{Local descriptors}
Local descriptors are detected by the Hessian-affine detector~\cite{mikolajczyk_scale_2004} and described by the SIFT local descriptor~\cite{SIFT_Lowe}. %We used same descriptors provided in~\cite{DBLP:conf/cvpr/JegouZ14}. 
RootSIFT variant~\cite{DBLP:conf/cvpr/ArandjelovicZ12} is used in all our experiments. 
For VLAT, FAemb, F-FAemb, at beginning, the SIFT descriptors
are reduced from 128 to 45 dimensions using PCA. For experiments with Fisher Vector, follow \cite{jegou11d}, we reduce SIFT descriptors to 64 dimensions using PCA. This makes the dimension of VLAT, FAemb, and F-FAemb comparable to dimension of compared embedding methods. 
%The training set to learn PCA projection for INRIA holidays and Oxford5k come from Flickr60k and Paris6k, respectively.

\subsubsection{Whitening and aggregating the embedded vectors}
%\subsection{Whitening and aggregating embedded vectors to single vector }
%\label{subsec:post}
%\subsubsection{Whitening}
%\label{subsubsec:post_white}
%In~\cite{DBLP:conf/cvpr/ChumM10}, authors showed that some local descriptors of an image are likely co-occurrence. The detector may also introduce some co-occurrence descriptors, for examples, when an image patch is described multiple times with different orientations~\cite{}, producing strongly co-occurrencing local descriptors. This phenomenon will effect the measure of similarity between two images. 

\textbf{Whitening}
Successful instance embedding methods consist of several feature post-processing steps. In~\cite{DBLP:conf/cvpr/JegouZ14}, authors showed that by applying the whitening processing, the discriminative power of embedded vectors can be improved, hence improving the retrieval results. %In addition, when dimension of $\phi(\x) \in \R^D$ is large, it may be redundant so we may need to reduce its dimension from $D$ to $D'$. In this case, the whitening operation can be performed jointly with dimensionality reduction. 
%They also suggested that in order to improve the localization of embedded vectors, we can remove first eigenvector associated with the largest eigenvalue.
In particular, given $\phi(\x) \in \R^{D}$, we achieve whitened embedded vectors $\phi_w(\x)$ by
\begin{equation}
\phi_w(\x) = diag\left( \lambda_1^{-\frac{1}{2}},\dots,\lambda_{D}^{-\frac{1}{2}} \right)P^T \phi(\x)
\end{equation}
%\begin{equation}
%\phi_w(\x) = diag\left( \lambda_2^{-\frac{1}{2}},\dots,\lambda_{D}^{-\frac{1}{2}} \right)P^T \phi(\x)
%\end{equation}
where $\lambda_i$ is the $i^{th}$ largest eigenvalue. $P \in \R^{D\times D}$ is matrix formed by the largest eigenvectors associated with the largest eigenvalues
%(ignoring largest eigenvectors) 
of the covariance matrix computed from learning embedded vectors $\phi(\x)$.

%It is worth to note that the whitening is part of Temb method. 
%Furthermore, Temb's authors~\cite{DBLP:conf/cvpr/JegouZ14} suggested that in order to improve the localization of embedded vectors, we can remove some first components associated with the largest eigenvalues. Following this suggestion, %However, by empirical experiments (with source code given by authors), we found that this operation is not necessary increasing the retrieval quality.
In \cite{DBLP:conf/cvpr/JegouZ14}, the authors further indicated that by discarding some first components associated with the largest eigenvalues of $\phi_w(\x)$, the localization of whitened embedded vectors will be improved. We apply this truncation operation in our experiments. The setting of this truncation operation is detailed in Section~\ref{subsec:compare_4methods}.%
%in order to improve the localization of embedded vectors, we can discard some first components associated with the largest eigenvalues. 
%In experiments, for Temb and VLAD methods, we discard $d$ first components of $\phi_w(\x)$ (as suggestion in~\cite{DBLP:conf/cvpr/JegouZ14}). The final dimension of $\phi_w(\x)$ is therefore $D = (n-1)\times d$. For VLAT and FAemb methods, we discard $\frac{d\times(d+1)}{2}$ first components of $\phi_w(\x)$. The final dimension of $\phi_w(\x)$ is therefor $D = \frac{(n-1) d (d+1)}{2}$.
\\
\\
\textbf{Aggregating} 
%\label{subsubsec:post_agg}
Let $\mathcal{X} = \{\x\}$ be set of local descriptors describing the image. Sum-pooling~\cite{DBLP:conf/cvpr/LazebnikSP06} and max-pooling~\cite{DBLP:conf/cvpr/YangYGH09,DBLP:conf/icml/BoureauPL10} are two common methods for aggregating set of whitened embedded vectors $\phi_w(\x)$ of the image to a single vector. Sum-pooling lacks discriminability because the aggregated vector is more influenced by frequently-occurring uninformative descriptors than rarely-occurring informative ones. Max-pooling equalizes the influence of frequent and rare descriptors. However, classical max-pooling approaches can only be applied to BoW or sparse coding features. 
Recently, in \cite{DBLP:conf/cvpr/JegouZ14}, the authors introduced a new aggregating method named~\textit{democratic aggregation} applied to image retrieval problem. This method bears similarity to generalized max-pooling~\cite{DBLP:conf/cvpr/MurrayP14} applied to image classification problem. Democratic aggregation can be applied to general features such as VLAD, Fisher vector\cite{jegou11d}, Temb\cite{DBLP:conf/cvpr/JegouZ14}. The authors \cite{DBLP:conf/cvpr/JegouZ14} showed that democratic aggregation achieves better performance than sum-pooling. The main idea of democratic aggregation is to find a weight for each $\phi_w(\x)$ such that
$\forall \x_i \in \mathcal{X}$
\begin{equation}
\lambda_i \left(\phi_w(\x_i)\right)^T \sum_{\x_j \in \mathcal{X}} \lambda_j \phi_w(\x_j) = 1
\end{equation}

In summary, the process for producing the single vector from the set of local descriptors describing the image is as follows. 
First, we map each $\x \in \mathcal{X} \to \phi(\x)$ and whiten $\phi(\x)$, producing $\phi_w(\x)$. We then use democratic aggregation to aggregate vectors $\phi_w(\x)$ to the single vector $\psi$ by
\begin{equation}
\psi(\mathcal{X}) = \sum_{\x_i \in \mathcal{X}} \lambda_i \phi_w(\x_i)
\end{equation}
\subsubsection{Power-law normalization}
The burstiness visual elements~\cite{DBLP:conf/cvpr/JegouDS09}, i.e., numerous descriptors almost similar within the same image, strongly affects the measure of similarity between two images. In order to reduce the effect of burstiness, we follow the common practical setting~\cite{herve_cvpr2010, DBLP:conf/cvpr/JegouZ14}: applying power-law normalization~\cite{DBLP:conf/eccv/PerronninSM10} to the final image representation $\psi$ and subsequently $L_2$ normalize it. The power-law normalization is applied to each component $a$ of $\psi$ by $a := |a|^{\alpha}sign(a)$, where $ 0 \le \alpha \le 1$ is a constant. We standardly set $\alpha = 0.5$ in our experiments.

\subsubsection{Rotation normalization and dimension reduction}
The power-law normalization suppresses visual burstiness but not the frequent
co-occurrences~\cite{DBLP:conf/eccv/JegouC12} that also corrupts the similarity measure. 
%In order to reduce the effect of co-occurrences, we follow~\cite{DBLP:conf/cbmi/SafadiQ13, DBLP:conf/cvpr/JegouZ14}, i.e., rotating data with a PCA rotation matrix learned on aggregated vectors from the learning set. 
In order to reduce the effect of co-occurrences, we follow~\cite{DBLP:conf/eccv/JegouC12,DBLP:conf/cvpr/JegouZ14}, i.e., rotating data with a whitening matrix learned on aggregated vectors from the learning set. 
%We use the same number of learning images as~\cite{DBLP:conf/cvpr/JegouZ14}. They are 10k images from Flickr60k for Holidays and Holidays+Flickr1M; 6k images from Paris6k for Oxford5k and Oxford105k. 
The results with the applying of this rotation are noted as +RN. When evaluating the short representations, we keep only first %$D'$ 
components, after RN, of aggregated vectors. 

\subsection{Impact of parameters and comparison between embedding methods}
\label{subsec:compare_4methods}
In this section, we compare FAemb, F-FAemb to other state-the-of-the art methods including VLAD~\cite{jegou11d}, Fisher \cite{jegou11d,DBLP:conf/cvpr/PerronninD07}, Temb~\cite{DBLP:conf/cvpr/JegouZ14} and VLAT~\cite{DBLP:conf/icip/PicardG11} under same test bed, i.e., the whitening, the democratic aggregation, and the power-law normalization are applied for all five embedding methods. 
\red{
We reimplement VLAD, VLAT in our framework. For Fisher, we use VLFeat library \cite{VLfeat}. It is worth noting that, in the design of these methods, the embedding and sum aggregating is combined into one formulation. Hence to apply the whitening and the democratic aggregation, we first apply VLAD/VLAT/Fisher embedding for each local feature separately. We then apply whitening and democratic aggregation on set of embedded vectors as usual. 
}
For Temb~\cite{DBLP:conf/cvpr/JegouZ14}, we use the implementation provided by the authors. 

Follow the suggestion in~\cite{DBLP:conf/cvpr/JegouZ14}, for Temb and VLAD methods, we discard first $d$(=128) components of $\phi_w(\x)$.  The final dimension of $\phi_w(\x)$ is therefore $D = (n-1)\times d$. 
\red{
For Fisher, we discard first $128$ components of $\phi_w(\x)$; this makes the dimension of Fisher equals to the dimension of VLAD and Temb.
} 
For VLAT, FAemb and F-FAemb methods, we discard first $\frac{d\times(d+1)}{2}$  components of $\phi_w(\x)$. The final dimension of $\phi_w(\x)$ is therefore $D = \frac{(n-1) d (d+1)}{2}$. The value of $\mu$ in (\ref{eq:objective}) and (\ref{eq:objective-ffaemb}) is selected by empirical experiments and is fixed to $10^{-2}$ for all FAemb, F-FAemb results reported bellow.

\red{The comparison between the implementation of VLAD, VLAT and Fisher in this paper and 
their improved versions~\cite{jegou11d, DBLP:journals/ieeemm/NegrelPG13} on Holidays dataset is presented in Table~\ref{tab:vlad_vlat}. It is worth noting that even with a lower dimension, the implementation of VLAD/VLAT/Fisher in our framework (RootSIFT descriptors, VLAD/VLAT/Fisher embedding, whitening, democratic aggregation and power-law normalization) achieves better retrieval results than %latest results 
their improved versions reported by the authors~\cite{jegou11d, DBLP:journals/ieeemm/NegrelPG13}.}

\begin{table}[!t]
\centering
\caption{The comparison between the implementation of VLAD, VLAT, Fisher in this paper and %the author's results 
their improved versions~\cite{jegou11d, DBLP:journals/ieeemm/NegrelPG13} on Holidays dataset. $D$ is the final dimension of aggregated vectors. Reference results are obtained from corresponding papers.}
\label{tab:vlad_vlat}
\begin{tabular}{|c|c|c|}
\hline
method	 	 	&$D$ 	 &mAP\\ \hline
VLAD~\cite{jegou11d}	&16,384	 &58.7 \\
VLAD (this paper)		&8,064	 &67.4 \\
VLAD (this paper)		&16,256	 &68.3 \\
\red{Fisher\cite{jegou11d}}   &\red{16,384}  &\red{62.5} \\
\red{Fisher (this paper)}     &\red{8,064}	 &\red{68.2}	\\
\red{Fisher (this paper)}     &\red{16,256}	 &\red{69.3}	\\
VLAT\textsubscript{improved}~\cite{DBLP:journals/ieeemm/NegrelPG13} 	 &9,000	&70.0  	\\ 
VLAT (this paper) 	 	&7,245	 &70.9  			\\ 
VLAT (this paper) 	 	&15,525	 &72.7  			\\\hline
\end{tabular}
\end{table}

\begin{figure}[!t]
\centering
\includegraphics[scale=0.5]{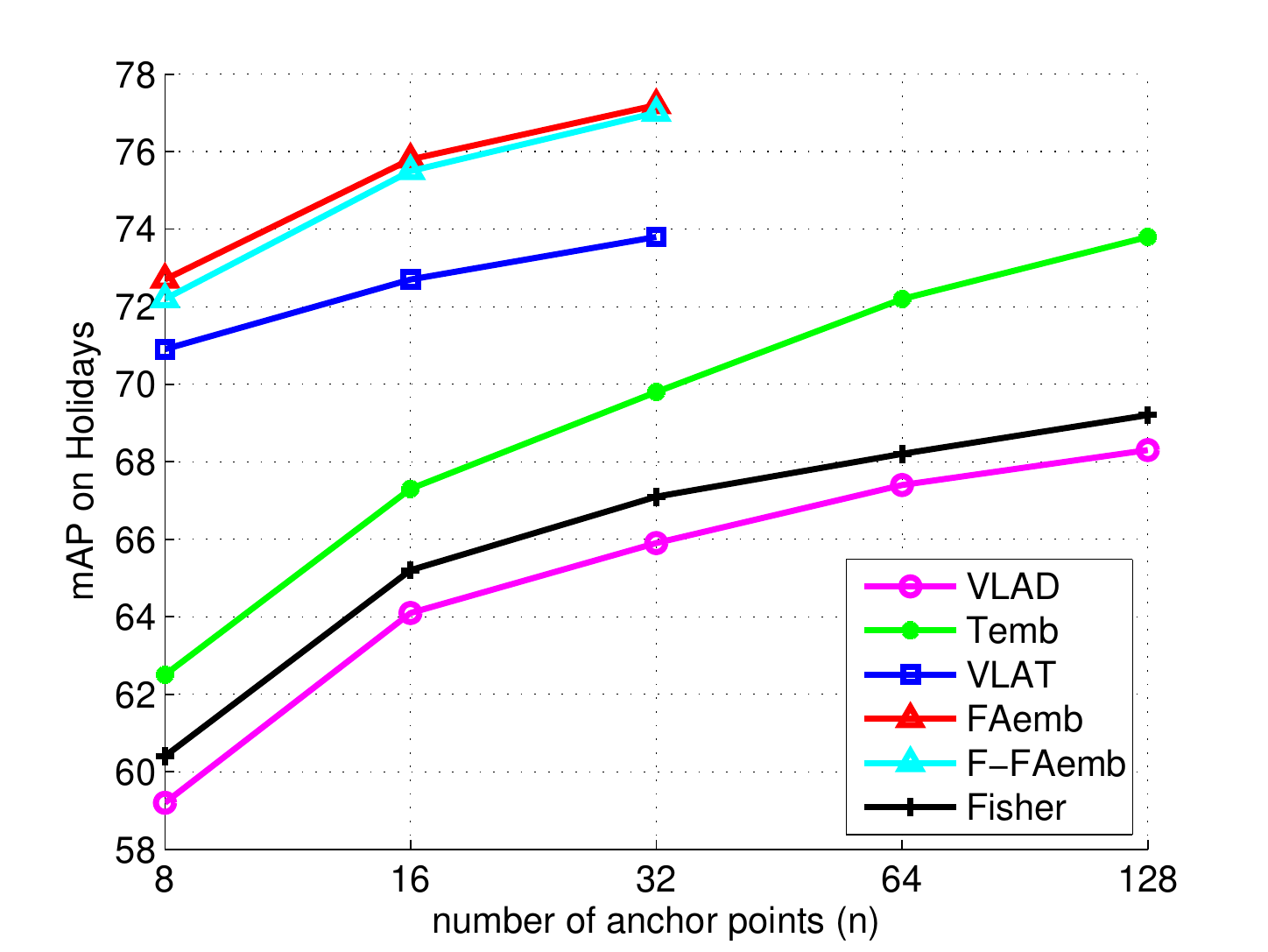}
\caption{Impact of number of anchor points on the Holidays dataset for different embedding methods: VLAD, Fisher, Temb, VLAT and the proposed FAemb, F-FAemb. Given $n$, the dimension of VLAD, Temb, Fisher is $128\times(n-1)$; the dimension of VLAT, FAemb, and F-FAemb is $\frac{45\times46}{2}\times (n-1)$.
}
\label{fig:map_holidays}
\end{figure}

\begin{figure}[!t]
\centering
\includegraphics[scale=0.5]{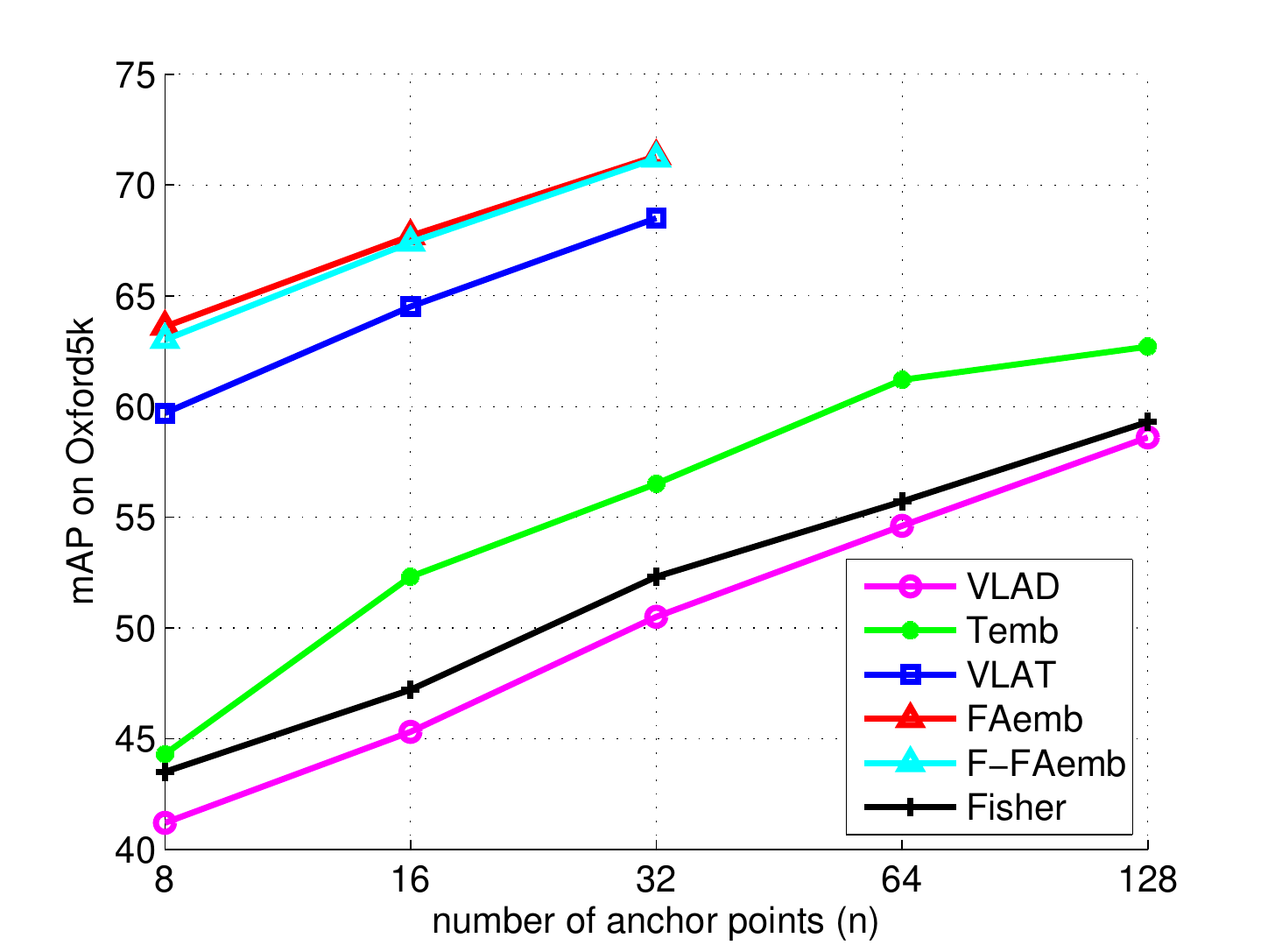}
\caption{Impact of number of anchor points on the Oxford5k dataset for different embedding methods: VLAD, Fisher, Temb, VLAT and the proposed FAemb, F-FAemb. Given $n$, the dimension of VLAD, Temb, Fisher is $128\times(n-1)$; the dimension of VLAT, FAemb, and F-FAemb is $\frac{45\times46}{2}\times (n-1)$.
}
\label{fig:map_oxford5k}
\end{figure}

\subsubsection{Impact of parameters} 
The main parameter here is the number of anchor points $n$. The analysis for this parameter is shown in Fig.~\ref{fig:map_holidays} and Fig.~\ref{fig:map_oxford5k} for Holidays and Oxford5k datasets, respectively. We can see that the mAP increases with the increasing of $n$ for all four methods. For all methods, the improvement tends to be smaller for larger $n$. This phenomenon has been discussed in~\cite{DBLP:conf/cvpr/JegouZ14}. For larger vocabularies, the interaction between descriptors is less important than for small ones. For VLAT, FAemb and F-FAemb, we do not report the results for $n>32$ as with this setting, the democratic aggregation is very time consuming. It has been indicated in~\cite{DBLP:conf/cvpr/JegouZ14} that when dimension of the embedded vector is high, e.g.  $> 32,000$, the benefit of democratic aggregation is not worth the computational overhead.

\subsubsection{Comparison between embedding methods} 
%It is worth noting that although the function bound of F-FAemb is a relaxed function bound of FAemb, F-FAemb provides a optimum solution on the coefficient vector while FAemb does not. This explains for the results that at same $n$, the performance F-FAemb is very competitive to FAemb. 
We find the following observations are consistent on both Holidays and Oxford5k datasets. 

The mAP of FAemb is slightly better than the mAP of F-FAemb at small $n$, i.e., $n=8$. When $n$ is large, i.e. $n=32$, F-FAemb and FAemb achieve very competitive results. 

At same $n$, FAemb, F-FAemb, and VLAT have same dimension. However, FAemb and F-FAemb improve the mAP over VLAT by a fair margin. For examples, at $n=8$, the improvement of FAemb over VLAT is \textbf{+1.8\%} and \textbf{+3.9\%} on Holidays and Oxford5k, respectively. At $n=16, 32$, the improvement is about \textbf{+3\%} on both datasets.  

{At comparable dimensions, FAemb and F-FAemb significantly improve the mAP over VLAD, Temb, Fisher.} 
For examples, comparing FAemb at ($n = 16, D = 15,525$) with VLAD/Temb at ($n = 128, D = 16,256$), the gain of FAemb over VLAD/Temb is \textbf{+7.5\%}/\textbf{+2\%} on Holidays and \textbf{+8.1\%}/\textbf{+5\%} on Oxford5k.

\subsection{Comparison with the state of the art}
\label{subsec:compare_soa}
In this section, we compare our framework %(RootSIFT descriptors, FAemb embedding, whitening, democratic aggregation, power-law normalization) 
with benchmarks having similar representation, i.e., they represent an image by a single vector. Due to the efficient computation of F-FAemb, it not only allows F-FAemb to use more anchor points for the function approximation but also allows F-FAemb to work on large scale datasets. Thus, we put more interest on F-FAemb when comparing to the state of the art. 
The main differences between the compared embedding methods are shown in Table~\ref{tab:benchmark}.

In VLAT\textsubscript{improved}~\cite{DBLP:journals/ieeemm/NegrelPG13}, VLAD\textsubscript{LCS}~\cite{DBLP:conf/mm/DelhumeauGJP13} and CVLAD~\cite{zhao2013oriented}, PCA and sum pooling are applied on Voronoi cells separately. Then, pooled vectors are concatenated to produce the single representation. 
In addition to methods listed in Table~\ref{tab:benchmark}, we also compare with the recent embedding method VLAD\textsubscript{LCS}+Exemplar SVM (VLAD\textsubscript{LCS}+ESVM)~\cite{Zepeda_2015_CVPR} and Convolutional Neural Network features. We consider the recent work~\cite{babenko2014neural} as the baseline for CNN-based image retrieval.  In~\cite{Zepeda_2015_CVPR}, the authors use the exemplar SVMs (linear SVMs trained with one positive example only and a vast collection of negative examples) as encoders. For each image, its VLAD\textsubscript{LCS}~\cite{DBLP:conf/mm/DelhumeauGJP13} representation is used as positive example for training an exemplar SVM. The weight vector (hyperplane) of the trained exemplar SVM is used as new representation. In~\cite{babenko2014neural}, the authors use the deep Convolutional Neural Network (CNN) model proposed in \cite{krizhevsky2012imagenet} for extracting image presentation. The network is first trained %using a subset ($1.2$ million images of $1,000$ classes) of 
on ImageNet dataset~\cite{ILSVRC15}. It is then retrained on the Landmarks dataset~\cite{babenko2014neural} containing $\sim 213,000$ images which are more relevant to the Holidays and the Oxford5k datasets. The activation values invoked by an image within top layers of the network are used as the image representation. It is worth noting that training CNN~\cite{babenko2014neural} is a supervised training task coming with challenges including: (i) the requirement for large amounts of labeled training data. %, i.e., the CNN~\cite{babenko2014neural} is first trained on ImageNet \cite{ILSVRC15} and is then retrained on $~213,000$ labeled Landmarks images. 
 According to~\cite{babenko2014neural}, the collecting of the labeled Landmarks images is a nontrivial task; (ii) the high computational cost and the requiring of GPUs. Contrary to CNN~\cite{babenko2014neural}, the training for our embedding is totally unsupervised, requiring of only several ten thousands of unlabeled images and without requiring of GPUs.
\red{ It is also worth noting that in \cite{babenko2014neural}, when evaluating on Holidays dataset, the authors rotate all images in the dataset to the normal orientation; when evaluating on Oxford5k dataset, they use the full queries, instead of using the cropped queries.  Both of these improve their results. In our experiments, we follow the literature \cite{jegou11d,DBLP:conf/cvpr/ArandjelovicZ13,DBLP:conf/mm/DelhumeauGJP13,DBLP:conf/cvpr/JegouZ14}, i.e., using the original images for Holidays and cropped queries for Oxford5k.}
\begin{table}[!t]
\footnotesize
\centering
\caption{The difference between compared embedding methods. RSIFT means RootSIFT. The second column indicates if PCA/whitening is applied on embedded vectors. The third column indicates aggregation method.}
\label{tab:benchmark}
\begin{tabular}{|c|c|c|c|}
\hline
Method	&Local  	 	&Do PCA/ 		&Aggr.	\\ 
     	&desc.		&whitening?		&method	\\ \hline
BoW~\cite{jegou11d}	&SIFT	&No	&Sum	 \\ 
VLAD~\cite{jegou11d}&SIFT	&No	&Sum	 \\ 
Fisher~\cite{jegou11d}&SIFT	&No	&Sum	 \\
VLAD\textsubscript{intra}~\cite{DBLP:conf/cvpr/ArandjelovicZ13} &RSIFT &No &Sum	 \\
VLAT\textsubscript{improved}~\cite{DBLP:journals/ieeemm/NegrelPG13} &SIFT	&PCA &Sum \\
%Fisher$\otimes$~\cite{tolias2014orientation}&RSIFT	&No		&Sum	\\
%VLAD$\otimes$~\cite{tolias2014orientation}	&RSIFT	&No		&Sum	\\
VLAD\textsubscript{LCS}~\cite{DBLP:conf/mm/DelhumeauGJP13} &RSIFT &PCA &Sum	\\
CVLAD~\cite{zhao2013oriented}	   &RSIFT	&PCA	&Sum\\
Temb~\cite{DBLP:conf/cvpr/JegouZ14} 		&RSIFT	&Whitening	&Democratic \\
FAemb										&RSIFT	&Whitening	&Democratic \\ 
F-FAemb										&RSIFT	&Whitening	&Democratic \\ \hline
\end{tabular}
\end{table}

\subsubsection{Evaluation on Holidays and Oxford5k datasets}
\begin{table}[!tb]
\small
\centering
\caption{Comparison with the state of the art on Holidays and Oxford5k datasets. $n$ is the  number of anchor points. $D$ is the dimension of the embedded vectors. %Reference results are obtained from corresponding papers.
} 
\label{tab:sota}
\begin{tabular}{|c| c| c|c|c|}
\hline
Method	&$n$ 	&$D$ 	 &\multicolumn{2}{|c|}{mAP}\\
    	&       & 			 &Hol. &Ox5k	\\ \hline
BoW~\cite{jegou11d} 	&200k 	 &200k  	&54.0	 &36.4 \\
%VLAD~\cite{jegou11d}	&64		 &4096	    &55.6	 &37.8 \\
VLAD~\cite{jegou11d}	&128	 &8,192   	&55.6	 &37.8 \\%From revisiting VLAD paper
VLAD~\cite{jegou11d}	&256	 &16,384   	&58.7	 &- \\
%Fisher~\cite{jegou11d}  &128	 &8,192 	&60.5    &41.8 \\%From revisiting VLAD paper
Fisher~\cite{jegou11d}  &256	 &16,384 	&62.5    &- \\
VLAD\textsubscript{LCS}~\cite{DBLP:conf/mm/DelhumeauGJP13}      &64  	&8,192 	&65.8	&51.7\\
VLAD\textsubscript{LCS}+ESVM~\cite{Zepeda_2015_CVPR}			&64		&8,192  &78.3 	&57.5\\
VLAD\textsubscript{intra}~\cite{DBLP:conf/cvpr/ArandjelovicZ13} &64  	&8,192  &56.5	&44.8\\
VLAD\textsubscript{intra}~\cite{DBLP:conf/cvpr/ArandjelovicZ13} &256 	&32,536 &65.3	&55.8 \\
CVLAD~\cite{zhao2013oriented}								&32		&32,768	&68.8	&42.7\\
VLAT\textsubscript{improved}~\cite{DBLP:journals/ieeemm/NegrelPG13} 	 &64  	&9,000	&70.0  	&- \\
%VLAD$\otimes$~\cite{tolias2014orientation}	
%Fisher$\otimes$~\cite{tolias2014orientation}	&64	&35,840	&\textbf{84.1}	&64.8 \\	
CNN~\cite{babenko2014neural}			&-	&4,096	&\textbf{79.3}	&54.5 \\
Temb~\cite{DBLP:conf/cvpr/JegouZ14}	 &64	&8,064  &72.2   &61.2 \\
Temb~\cite{DBLP:conf/cvpr/JegouZ14}	 &128	&16,256 &73.8   &62.7 \\ 
FAemb				 &8		&7,245	&72.7	&63.6 \\
FAemb				 &16	&15,525	&75.8	&67.7 \\
%FAemb~\cite{do_cvpr15}				 &32	&32,085	&77.2	&\textbf{71.3} \\ \hline
%\multicolumn{5}{|c|}{Our framework} \\ \hline
F-FAemb							 &8		&7,245	&72.2	&63.4\\
F-FAemb						 &16	&15,525	&75.5	&67.6\\
F-FAemb							 &32	&32,085	&{77.0}	&\textbf{70.7}\\ \hline
\multicolumn{5}{|c|}{With rotation normalization} \\ \hline
Temb +RN~\cite{DBLP:conf/cvpr/JegouZ14}		&64		&8,064	&77.1	&67.6 \\
Temb +RN~\cite{DBLP:conf/cvpr/JegouZ14}		&128	&16,256	&76.8	&66.5 \\
FAemb +RN							&8		&7,245	&76.2   &66.7 \\ 
FAemb +RN							&16		&15,525	&78.7   &70.9 \\ 
%FAemb + RN										 &32	&32,085	&-	&- \\ 
F-FAemb +RN						    &8		&7,245	&75.5   &66.1 \\ 
F-FAemb +RN							&16		&15,525	&78.6   &70.3 \\ 
F-FAemb +RN							&32		&32,085	&\textbf{80.7}	&\textbf{74.2} \\ \hline
\end{tabular}
\end{table}
Table~\ref{tab:sota} shows the results of FAemb, F-FAemb, and the compared methods on Holidays and Oxford5k datasets. %The first observation is that by achieving an efficient computation, F-FAemb allows the use of more anchor points (i.e. $n=32$) for embedding, and this boosts the performance.

Without RN post-processing, F-FAemb outperforms or is competitive to most compared methods. CNN features \cite{babenko2014neural} achieve best performance on the Holidays dataset; its mAP is higher than F-FAemb ($n=32$) $+2.3\%$. However, on the Oxford5k dataset, F-FAemb outperforms CNN features~\cite{babenko2014neural} by a fair margin, i.e., $+16.2\%$.

When RN is used, it boosts performance for all Temb, FAemb and F-FAemb. The performance of F-FAemb +RN at $D=7,245$ is slightly lower than Temb+RN at $D=8,064$. However, at higher dimension, i.e., $D=15,525$, the performance of F-FAemb + RN outperforms all performances of Temb+RN a fair margin. 

The efficient computation of F-FAemb allows it to use high number of anchor points, i.e., $n=32$; and at this setting, F-FAemb +RN outperforms all compared methods on both datasets. The gain is more significant on the Oxford5k dataset, i.e., F-FAemb +RN outperforms the recent embedding method VLAD\textsubscript{LCS}+ESVM~\cite{Zepeda_2015_CVPR} $+16.7\%$ and outperforms the CNN features~\cite{babenko2014neural} $+19.7\%$. It is worth noting that the dimension of the CNN features is lower than ones of F-FAemb +RN. We evaluate the performance of F-FAemb +RN in case of short representation in Section~\ref{subsubsec:short_rep}. 
\red{It is worth noting that in \cite{DBLP:conf/icip/NegrelPG14}, the authors report strong results on Holidays dataset, i.e., mAP = 84. However, in that work, to describe an image patch, they use two types of local descriptors: HOG feature \cite{DBLP:conf/cvpr/DalalT05} and color feature \cite{DBLP:conf/eccv/PerronninSM10}. They apply VLAT for each type of descriptor separately, and concatenate two resulted embedded vectors, producing the single representation of 1.7M dimension. Hence, that work may not directly compare to ours and other works in Table \ref{tab:benchmark} in which only SIFT feature is used.} 
\subsubsection{Evaluation on large scale dataset: Oxford105k}
\begin{table}[!h]
\centering
\caption{Comparison with the state of the art on Oxford105k dataset. $D$ is dimension of the  embedded vectors. %Reference results are obtained from corresponding papers.
} 
\label{tab:sota_105k}
\small
\begin{tabular}{|c|c|  c|}
\hline
Method 	 	&$D$ 	 &\multicolumn{1}{|c|}{mAP}\\
     		&		&Oxford105k		\\ \hline
VLAD\textsubscript{LCS}~\cite{DBLP:conf/mm/DelhumeauGJP13}      &8,192 	&45.6  \\
%VLAD$\otimes$~\cite{tolias2014orientation}	    				&28,768 &53.9  \\
Temb+RN~\cite{DBLP:conf/cvpr/JegouZ14}	 	&8,064  &61.1   \\
CNN~\cite{babenko2014neural}		&4,096  &51.2	\\		
%FAST-FAemb									&7,245	&62.6	&\\	 		
F-FAemb +RN							 	&7,245	&64.3	\\
%FAST-FAemb									&15,525 &67.3 	&\\
F-FAemb +RN							 	&15,525	&\textbf{68.1}	\\ \hline
\end{tabular}
\end{table}
%To evaluate the scale-up ability of F-FAemb, we experiment on the Oxford105k dataset.
The Oxford105k dataset is used for large scale testing in a few benchmarks~\cite{DBLP:conf/cvpr/JegouZ14, DBLP:conf/mm/DelhumeauGJP13,babenko2014neural}. %Because FAemb~\cite{do_cvpr15} is time consuming to compute the coefficients, it is difficult to apply to large scale dataset.~\cite{do_cvpr15} does not report results on Oxford105k dataset. 
The comparative mAP between methods is shown in Table~\ref{tab:sota_105k}. The results show that even with a lower dimension, the proposed F-FAemb (at $D=7,245$) outperforms the compared methods (VLAD\textsubscript{LCS}, Temb+RN) by a large margin. The best result of F-FAemb (i.e. at $D=15,525$) sets up state-of-the-art performance on this large scale dataset. It outperforms the current state of the art, i.e., Temb+RN~\cite{DBLP:conf/cvpr/JegouZ14} $+7\%$.

\subsubsection{Evaluation on short representation}
%% for running Oxford105k
% TTD/OriginalFromQuang/FastFaemb_L1_Final_5 ---> xem file img_eval_trunc.m
% The result for short representation is also available at
% D:\Dropbox\MINDEF_report\iccv2015_L1\figures\FastFA_LLC_results
% Quang prepared this for ICCV'15
\label{subsubsec:short_rep}
\begin{table}[!h]
\small
\centering
\caption{Comparison with the state of the art on short vector representation. The original dimension of F-FAemb +RN is $7,245$. The original dimension of Temb+RN on Holidays, Oxford5k, Oxford105k is $8,064$ and on Holidays+Flickr1M (Hol.+Fl1M) is $1,920$. %The original dimension is reduced to $D'$ by PCA. 
} 
\label{tab:sota_short}
\begin{tabular}{|c|c| c c c c|}
\hline
Method		&$D$ 	 &\multicolumn{4}{|c|}{mAP}\\
				&		 &Hol. 	&Ox5k	&Ox105k  &Hol.+Fl1M\\ \hline
%				&		 &		&		&		 &Fl1M	\\ 
%VLAD$\otimes$~\cite{tolias2014orientation}&1,024	&- &- &40.7 \\
Temb+RN~\cite{DBLP:conf/cvpr/JegouZ14}	  &1,024 &\textbf{72.0} &56.2 &50.2   &49.4\\
F-FAemb +RN					&1,024 &70.8 &\textbf{58.2} &\textbf{53.2} &\textbf{68.5} \\ 
\hline
Temb+RN~\cite{DBLP:conf/cvpr/JegouZ14}	  &512 	 &\textbf{70.0} &52.8 &46.1   &46.9\\
F-FAemb +RN					&512   &69.0 &\textbf{53.9} &\textbf{50.9} &\textbf{65.3}\\
\hline
{Temb+RN~\cite{DBLP:conf/cvpr/JegouZ14}}	  &{256} 	 &{65.7}  &{\textbf{47.2}}   &{40.8}   &{43.7}\\
\red{F-FAemb +RN}					     &\red{256}   &\red{\textbf{67.5}} &\red{45.6} &\red{\textbf{44.5}} &\red{\textbf{61.9}}\\
\hline
Temb+RN~\cite{DBLP:conf/cvpr/JegouZ14}	  &128 	 &61.5   &\textbf{40.0}    &33.9   &38.7\\
%VLAD\textsubscript{intra}~\cite{DBLP:conf/cvpr/ArandjelovicZ13} &128 &62.5 &44.8 &37.4 &-\\
VLAD\textsubscript{LCS}~\cite{DBLP:conf/mm/DelhumeauGJP13}  &128 &- &32.2 &26.2 &39.2\\
\red{F-FAemb +RN}					&\red{128}   &\red{\textbf{63.0}} &\red{{39.4}} &\red{\textbf{37.1}} &\red{\textbf{58.0}}\\
\hline

CNN~\cite{babenko2014neural}	  &4,096 &\textbf{79.3}	&54.5	&51.2	&-\\		
%FAST-FAemb									&7,245	&62.6	\\	 		
%FAST-FAemb+RN							 	&7,245	&64.3	\\
%FAST-FAemb									&15,525 &67.3 	\\
F-FAemb +RN					&4,096	&74.1 &\textbf{63.7} &\textbf{62.2}	&\textbf{72.5}\\ \hline
\end{tabular}
\end{table}
As the F-FAemb features are high-dimensional, a question of their performance on short representations arises. In this section, we evaluate the performance of F-FAemb at short representations achieved by keeping only first components after the rotation normalization of aggregated vectors.  Table~\ref{tab:sota_short} reports comparative mAP for varying dimensionality. %We compare our framework with the state of the art having same dimensionality. 

Compare to Temb+RN at the same dimension, on Holidays dataset, Temb+RN is slightly better than our method at D = 1024 and 512, while at D=256 and 128, F-FAemb+RN outperforms Temb+RN. 
On Oxford5k dataset,  our method outperforms Temb+RN at D = 1024 and 512l, while at D=256 and 128, Temb+RN is slightly better than ours. 
On large scale datasets Oxford105k and Holidays+Flickr1M, our method significantly improves the mAP over Temb+RN. On Oxford105k, the gains are $+3\%$, $+4.8\%$, $+3.7\%$ and $+3.2\%$ at $1024$, $512$, $256$ and $128$ dimensions, respectively. On Holidays+Flickr1M, the gains are $+19.1\%$, $+18.4\%$, $+18.2\%$, and $+19.3\%$ at $1024$, $512$, $256$, and $128$ dimensions, respectively.

Compare to CNN features having $4,096$ dimension, the performance of CNN features is higher than our method $+5.2\%$ on Holidays dataset, but we see much larger variances on Oxford5k and Oxford105k datasets. The gains of our method over CNN features on Oxford5k and Oxford105k are $+9.2\%$ and $+11\%$, respectively.
\subsubsection{Evaluation on binary representation}
Two main problems which need to be considered in large scale image search are fast searching and efficient storage. An attractive approach for handling those problems is to represent each image by very compact codes, i.e., binary codes %\cite{DBLP:journals/pami/GongLGP13,DBLP:conf/cvpr/HeWS13,DBLP:conf/eccv/DoDNC16,CVPR12:SphericalHashing,BA_CVPR15,UH-BDNN-Do2016,DBLP:journals/tmm/DuanLWHG15}. 
\cite{DBLP:journals/pami/GongLGP13,DBLP:conf/cvpr/HeWS13,BA_CVPR15,UH-BDNN-Do2016,DBLP:journals/tmm/DuanLWHG15}. 

 In this section, we further evaluate the performance of the proposed F-FAemb when the single representation  is compressed to compact binary codes. In order to achieve compact codes for the single representation, we use the state-of-the-art hashing method Iterative Quantization (ITQ)~\cite{DBLP:journals/pami/GongLGP13}. 
The ITQ has two main steps: the first step is to apply PCA for dimensionality reduction; the second step is to seek an optimal rotation matrix which rotates the projected data to binary values such that the quantization error is minimized. %As the RN post-processing is actually a PCA projection which is similar to the first step of ITQ, it is unnecessary to apply RN post-processing on the single representation before ITQ. We empirically observe that applying RN on the single representation before ITQ does not necessarily improve retrieval results.

We compare our F-FAemb to the recent embedding method Temb~\cite{DBLP:conf/cvpr/JegouZ14} when both of them are compressed to binary codes. 
%\begin{figure}[!t]
%	\centering
%	\subfigure[Holidays]{
%		\includegraphics[width=3.9cm,height=3.9cm]{binary_holidays_k8.pdf}
%		\label{fig:map_binary_holidays}
%	}
%	\subfigure[Oxford5k]{
%		\includegraphics[width=3.9cm,height=3.9cm]{binary_ox5k_k8.pdf}
%		\label{fig:map_binary_oxford5k}
%	}
%	
%	\subfigure[Oxford105k]{
%		\includegraphics[width=3.9cm,height=3.9cm]{binary_ox105k_k8.pdf}
%		\label{fig:map_binary_oxford105k}
%	}
%	\caption[]{Comparison between F-FAemb and Temb in binary representation, mAP in three datasets: Holidays, Oxford5k and Oxford105k using $128$ to $1024$ bits. }
%	\label{fig:map_binary}
%\end{figure}
\begin{table}
	\small
	\centering
	\caption{mAP comparison between F-FAemb and Temb in binary representation with varying code lengths on three datasets (Holidays, Oxford5k and Oxford105k). The original dimension of F-FAemb and Temb are $7,245$ and $8,064$, respectively.}
	\begin{tabular}{|l|c|c c c c|}
		
		\hline
		\multirow{2}{*}{Dataset} & \multirow{2}{*}{Method} & \multicolumn{4}{|c|}{Code length (bits)} \\
														  \cline{3-6} 
								 &						   & 128  & 256  & 512  & 1024 \\
		\hline \hline
		\multirow{2}{*}{Holidays}     & Temb\cite{DBLP:conf/cvpr/JegouZ14}                    & 39.2 & 46.5 & 53.0 & 57.3\\
								 & F-FAemb  & \textbf{40.1} & \textbf{47.9} & \textbf{54.5} & \textbf{59.7}\\
		\hline
		\multirow{2}{*}{Ox5k}    & Temb\cite{DBLP:conf/cvpr/JegouZ14}                    & \textbf{27.1} & 33.1 & 38.5 & 43.4\\
								 & F-FAemb            & 26.4 & \textbf{33.8} & \textbf{40.7} & \textbf{45.9}\\
		\hline
		\multirow{2}{*}{Ox105k}  & Temb\cite{DBLP:conf/cvpr/JegouZ14}                    & \textbf{25.9} & 31.6 & 37.7 & 42.9\\
								 & F-FAemb            & 24.2 & \textbf{32.0} & \textbf{38.5} & \textbf{44.7}\\
		\hline
	\end{tabular}
	\label{tab:map_binary}
	
\end{table}
The comparative results are presented in Table \ref{tab:map_binary}. On Holidays dataset, F-FAemb achieves better results than Temb for all code lengths; the improvement increases with the increasing of the code length. On Oxford5k and Oxford105k datasets, Temb is better than F-FAemb at low code length, i.e., 128-bit codes. However, F-FAemb outperforms Temb when the number of bits is increase, i.e. $>128$; the improvement is more clear at high code lengths. %Concretely, in Oxford5k, the gain in $256$, $512$ and $1024$ bits respectively are $0.6$\%, $2.1$\% and $2.4$\%. In Oxford105k, $0.3$\%, $0.8$\% and $1.8$\% are also the improvements of F-FAemb over Temb in $256$-, $512$- and $1024$-bit code length.

\subsection{Results when CNN are used as local features}
\label{subsec:cnn}
In this section we further evaluate the proposed F-FAemb when the image is described by a set of CNN features which are state-of-the-art image representation for various computer vision tasks \cite{DBLP:conf/cvpr/RazavianASC14}. 

\subsubsection{Configuration}
\label{subsub:config_CNN}
Specifically, instead of using set of local SIFT features to describe the image as previous experiments, we extract CNN activations for local patches at multiscale levels. We then take the union of all the patches from the image, regardless of scale. This union set can be considered as local features describing the image. We use the output of the last fully connected layer of the pretrained AlexNet model \cite{jia2014caffe} as CNN features representing for patches. We extract CNN activations at 3 levels. For the first level, we simply take 4096-dimensional CNN activations for the whole image. For the second and the third levels, we extract CNN activations for all $128\times 128$, $64\times64$ patches sampled with a stride of 30 pixels. In order to make the computation of the embedding more efficient, we use PCA to reduce 4096-dimensional features to 45-dimensional features. 
The same processing (i.e., F-FAemb embedding, whitening, democratic aggregating, rotation normalization, power normalization) is applied on the set of CNN features to produce the single representation.
\red{
\subsubsection{Comparison to the state of the art}
\label{subsub:compareSOTA_CNN}
We compare our CNN features-based F-FAemb with the state of the art which use Convolutional Neural Network, deep learning techniques for image retrieval, i.e., Convolutional Kernel Networks (CKN) \cite{CKN-paulin}, the combination of Fisher Vector encoding and Deep Neural Network (FV-DNN) \cite{DBLP:conf/cvpr/PerronninL15}, Multiscale Orderless Pooling (MOP-CNN) \cite{DBLP:conf/eccv/GongWGL14}, CNN features (CNN) \cite{DBLP:conf/cvpr/RazavianASC14}. We also compare F-FAemb to very recent works: Sum Pooling of Convolutional feature (SPoC) \cite{DBLP:conf/iccv/BabenkoL15}, Regional Maximum Activation of Convolutional feature (R-MAC) \cite{DBLP:journals/corr/ToliasSJ15}. 

Among mentioned works, FV-DNN \cite{DBLP:conf/cvpr/PerronninL15}, MOP-CNN \cite{DBLP:conf/eccv/GongWGL14}, CNN \cite{DBLP:conf/cvpr/RazavianASC14} rely on the outputs of a fully connected layer, while the recent SPoC \cite{DBLP:conf/iccv/BabenkoL15} and R-MAC \cite{DBLP:journals/corr/ToliasSJ15} apply the pooling (sum pooling or max pooling) on the activations of a convolutional layer for producing the single representation. It is also worth mentioning that the recent work \cite{DBLP:conf/cvpr/CimpoiMV15} applies the Fisher Vector encoding on the outputs of a convolutional layer. That work, however, evaluates on the texture recognition problem, not on image retrieval. 

Note that when evaluating on Oxford5k dataset, FV-DNN \cite{DBLP:conf/cvpr/PerronninL15}, CNN \cite{DBLP:conf/cvpr/RazavianASC14} report results with ``full query"; SPoC \cite{DBLP:conf/iccv/BabenkoL15}, CKN \cite{CKN-paulin} report results with both ``full query" and ``crop query", while R-MAC \cite{DBLP:journals/corr/ToliasSJ15} reports results of only ``crop query".
It is also worth noting that in \cite{DBLP:conf/cvpr/RazavianASC14}, at the retrieval stage on the Oxford5k, they use the spatial search, i.e.,  for each image they extract multiple patches of different sizes/scales and compute CNN representation for patches. The distance between a query patch and a reference image is defined as the minimum $l_2$ distance between the query patch and respective reference patches. The distance between the query and the reference image is set to the average distance of each query patch to the reference image. This spatial search is a costly searching and it differs from ours and \cite{CKN-paulin,DBLP:conf/cvpr/PerronninL15,DBLP:conf/eccv/GongWGL14,DBLP:conf/iccv/BabenkoL15,DBLP:journals/corr/ToliasSJ15} in which only a single representation is matched per image.

Aforementioned works apply PCA/whitening on the single representation. Thus, for clear presentation, we ignore the \textit{+RN} notation when presenting results of F-FAemb. The comparative mAP is showed in Table \ref{tab:soa_cnn}. The short vector representations of F-FAemb (i.e., when $D<7,245$) is achieved by keeping only first components after the rotation normalization step. %It is not surprised that when the CNN features, which is state-of-the-art image representation, is used for FAemb, it significantly boosts the mAP in comparison to the using of SIFT, e.g., at $D=7,245$ the mAP is improved $10.5\%$ and \textcolor{red}{xxx} for Holidays and Oxford5k, respectively. 

\begin{table}[!t]
\centering
\small
\caption{Comparison with state-of-the-art CNN / deep learning-based image retrieval.} 
\label{tab:sota_cnn}
\begin{tabular}{|c|c| c c c|}
\hline
Method		&$D$ 	 &\multicolumn{3}{|c|}{mAP}\\
			&		 &Hol. 	&Oxford5k			&Oxford5k\\ %&Ox105k  
			&		 &			&(full query)		&(crop query)\\ \hline
%F-FAemb										&15,525  &00.0  &00.0 \\
F-FAemb										&7,245   &\textbf{86.0}    &59.5 	&56.3\\\hline% \\\hline		
F-FAemb										&4,096   &{85.4}  &57.8 	&54.0\\ % \\ 
CKN\cite{CKN-paulin}						&4,096	 &82.9	  &55.4  	&56.5\\% 
FV-DNN\cite{DBLP:conf/cvpr/PerronninL15}	&4,096 	 &84.7    &-   		&-  \\
CNN\cite{DBLP:conf/cvpr/RazavianASC14}		&4,096 	 &84.3    &\textbf{68.0} 	&-\\
\hline
F-FAemb									    &2,048   &{83.5}  &53.4   	&50.7\\% \\ 
MOP-CNN\cite{DBLP:conf/eccv/GongWGL14}	    &2,048   &80.2    &- 	    &-	\\\hline

F-FAemb									    &512     &{78.1}  &41.2 	&39.3\\% \\ 
R-MAC\cite{DBLP:journals/corr/ToliasSJ15}	&512     &{-}     &- 		&\textbf{66.9}\\ \hline

F-FAemb									    &256     &{74.0}  &36.6 	&34.5\\% \\ 
R-MAC\cite{DBLP:journals/corr/ToliasSJ15}	&256     &{-}     &- 	    &56.1\\% \\ 
SPoC \cite{DBLP:conf/iccv/BabenkoL15}	    &256     &{80.2}  &58.9 	&53.1\\ \hline

\end{tabular}
\label{tab:soa_cnn}
\end{table}

%On Holidays dataset, at the same dimension, F-FAemb is on par to FV-DNN\cite{DBLP:conf/cvpr/PerronninL15}, CNN\cite{DBLP:conf/cvpr/RazavianASC14} and considerably improves over MOP-CNN\cite{DBLP:conf/eccv/GongWGL14}, CKN\cite{CKN-paulin}.  

On the Holidays dataset, at the same dimension, F-FAemb slightly improves over FV-DNN\cite{DBLP:conf/cvpr/PerronninL15}, CNN\cite{DBLP:conf/cvpr/RazavianASC14} and considerably improves over MOP-CNN\cite{DBLP:conf/eccv/GongWGL14}, CKN\cite{CKN-paulin}.
When short representation is applied, i.e. $D=256$, the recent SPoC \cite{DBLP:conf/iccv/BabenkoL15} outperforms F-FAemb. However, it is worth noting that F-FAemb achieves state-of-the-art results at its full dimension, i.e., mAP = $86$  at $D=7,245$.

On the Oxford5k dataset, when the ``full query" is used, F-FAemb outperforms CKN \cite{CKN-paulin} %\footnote{The results of CKN\cite{CKN-paulin} on Oxford5k is with cropped queries. According to \cite{CKN-paulin}, the cropped queries achieve better performance than full queries, i.e., the their best mAP with full queries is $55.4$.} 
while it is lower than CNN\cite{DBLP:conf/cvpr/RazavianASC14}. However it is worth noting that the spatial search used in \cite{DBLP:conf/cvpr/RazavianASC14} is a costly searching as it has two drawbacks. First, all the patch vectors of the image have to be stored. This increases the memory requirements by a factor of $P$ where $P$ is number of extracted patches per image. Second, the complexity for computing the distance between two images is increased by a factor of $P^2$. Contrary to \cite{DBLP:conf/cvpr/RazavianASC14}, F-FAemb and other methods only store a single representation per image and compute only one Euclidean distance when comparing two images. When the short representation is applied, SPoC \cite{DBLP:conf/iccv/BabenkoL15} outperforms F-FAemb. In \cite{DBLP:conf/iccv/BabenkoL15}, the authors show that the convolutional features is robust to PCA compression, i.e., applying PCA on the single representation improve the mAP rather than decrease it as other methods. 
When the ``crop query" is used, the recent R-MAC \cite{DBLP:journals/corr/ToliasSJ15} gives very strong results; it outperforms all compared methods. 

Compare F-FAemb + SIFT features (Table \ref{tab:sota_short}, Table \ref{tab:sota}) to F-FAemb + CNN features (Table \ref{tab:soa_cnn}), we have the observation that the used configuration of CNN features (Section \ref{subsub:config_CNN}) achieves better results than SIFT features on Holidays dataset which contains general images. However, when testing on Oxford5k dataset containing particular objects (buildings), the SIFT features achieve better performance than CNN features. In order to make easy comparison for other researches, we summarize our best results in Table \ref{tab:bestF-FAemb}, in which CNN features are used for Holidays dataset and SIFT features are used for Oxford5k dataset (crop query).

\begin{table}[!t]
\centering
\small
\caption{Best results of F-FAemb (with intermediate/short representation). CNN features are used for Holidays; SIFT features are used for Oxford5k} 
\label{tab:sota_cnn}
\begin{tabular}{|c|c|  c c|}
\hline
Method		&$D$ 	 &\multicolumn{2}{|c|}{mAP}\\
			&		 &Holidays 				&Oxford5k\\ %&Ox105k  
			&		 &					&(crop query)\\ \hline
%F-FAemb										&15,525  &86.8     	&70.3\\	
%F-FAemb										&7,245   &86.0     	&66.1\\	
%F-FAemb										&4,096   &85.4   	&63.7\\ % \\ 
%F-FAemb									    &1,024   &81.4     	&58.2\\% \\ 
%F-FAemb									    &512     &78.1  	&53.9\\% \\ 
%F-FAemb									    &256     &74.0   	&45.6\\\hline
\multirow{6}{*}{F-FAemb}									&15,525  &86.8     	&70.3\\	
										&7,245   &86.0     	&66.1\\	
										&4,096   &85.4   	&63.7\\ % \\ 
									    &1,024   &81.4     	&58.2\\% \\ 
									    &512     &78.1  	&53.9\\% \\ 
									    &256     &74.0   	&45.6\\\hline

\end{tabular}
\label{tab:bestF-FAemb}
\end{table}
}

\section{Conclusion}
\label{sec:concl}
Embedding local features to high dimensional space is a crucial step for producing the single powerful image representation in many state-of-the-art large scale image search systems. In this paper, by departing from the goal of linear approximation of a nonlinear function in high dimensional space, we first propose a novel embedding method. 
The proposed embedding method, FAemb, can be seen as the generalization of several well-known embedding methods such as VLAD, TLCC, VLAT. In order to speed up the embedding process, we then derive the fast version of FAemb, in which the embedded vector can be efficiently computed, i.e., in the closed-form. 
The proposed embedding methods are evaluated with different state-of-the-art local features such as SIFT, CNN, in image search context under various settings. The experimental results show that the proposed embedding methods give a performance boost over the state of the art on several standard public image retrieval benchmarks.

\appendices
\section{Proof of Lemma~\ref{lemma_1}}
\label{proof_lemma_1}
%First, let us introduce and proof following lemma
%\begin{lemma}
%If $\R^d \to \R$ is of class of $C^{k+1}$ and $\nabla^k f(\x)$ is Lipschitz continuous, then exist $M$ such that $|\partial^{\alpha}f(\x)| \le M$ for $\x\in R^d$ and $|\alpha| = k+1$
%\end{lemma}
%\textbf{Proof:} 
By assumption, we have (i) $f$ is of class of $C^{k+1}$. 
Because $\nabla^k f(\x)$ is Lipschitz continuous with constant $M>0$, we have $\norm{\nabla^{k+1} f(\x)}_{2} \le M$. So for $| \alpha | = k+1$, we have (ii) $\vert \partial^{\alpha} f( \x )\vert \le \norm{\nabla^{k+1} f(\x)}_{2} \le M$.
%\begin{equation}
% \vert \partial^{\alpha} f( \x )\vert \le \norm{\nabla^{k+1} f(\x)}_{2} \le M
% \label{eq:cond}
%\end{equation}
 (i) and (ii) make the condition of the Corollary \ref{bound_remainder} is held. 
%Because $\nabla^k f(\x)$ is Lipschitz continuous with constant $M$, we have $\norm{\nabla^{k+1} f(\x)}_{2} \le M$. This means that $\vert \partial^{\alpha} f( \x )\vert \le M$ for $| \alpha | = k+1$

We have
\begin{eqnarray}
{} & & \left| f(\x) - \sum_{j=1}^{n} \gamma_{\v_j}(\x)  \sum_{ | \alpha | \le k} \frac{\partial^{\alpha} f(\v_j)}{\alpha!} \left( \x -  \v_j \right)^\alpha \right|  \nonumber \\
{} & = & \left| \sum_{j=1}^{n} \gamma_{\v_j}(\x) \left( f(\x) -   \sum_{ | \alpha | \le k} \frac{\partial^{\alpha} f(\v_j)}{\alpha!} \left( \x -  \v_j \right)^\alpha \right) \right| \nonumber \\
{} & \le &  \sum_{j=1}^{n} \left| \gamma_{\v_j}(\x) \left( f(\x) -   \sum_{ | \alpha | \le k} \frac{\partial^{\alpha} f(\v_j)}{\alpha!} \left( \x -  \v_j \right)^\alpha \right) \right| \nonumber \\
{} & = &  \sum_{j=1}^{n} \left| \gamma_{\v_j}(\x) \right| \left|   R_{\v_j, k}(\x - \v_j)  \right| \nonumber \\
{} &\le& \frac{M}{(k+1)!} \sum_{j=1}^{n} \left|\gamma_{\v_j}(\x) \right| \norm{\x-\v_j}_{1}^{k+1} \nonumber .
\end{eqnarray}
where the last inequation comes from the Corollary~\ref{bound_remainder}.

%% use section* for acknowledgment
%\ifCLASSOPTIONcompsoc
%  % The Computer Society usually uses the plural form
%\else
%  % regular IEEE prefers the singular form
%\fi

\ifCLASSOPTIONcaptionsoff
  \newpage
\fi

{%\small
\bibliographystyle{IEEEtran}
\bibliography{FAemb}
}

%\end{IEEEbiography}
%% insert where needed to balance the two columns on the last page with
%%\newpage

\vspace{-0.5cm}
\begin{IEEEbiography}[{\includegraphics[width=1in,height=1.25in,clip,keepaspectratio]{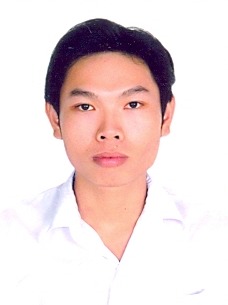}}]
{Thanh-Toan Do}
received the Ph.D. degree in computer science from the University of Rennes 1, France, in 2012. From 2009-2012, he did his Ph.D. at INRIA Rennes, France. He is
currently a Research Fellow with the University of Adelaide, Australia. From 2013-2016, he was a postdoctoral researcher with the Singapore University of Technology and Design (SUTD), Singapore. His research interests are on computer vision and machine learning.
\end{IEEEbiography}
\vspace{-0.6cm}
\begin{IEEEbiography}[{\includegraphics[width=1in,height=1.25in,clip,keepaspectratio]{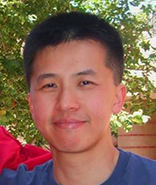}}]
{Ngai-Man Cheung}
received the Ph.D. degree in electrical engineering from the University  of Southern California, Los Angeles, CA, in 2008. He is currently an Assistant Professor with the Singapore University of Technology and Design (SUTD). From 2009-2011, he was a postdoctoral researcher with the Image, Video and Multimedia Systems group at Stanford University, Stanford, CA.  He has also held research positions with Texas Instruments Research Center Japan, Nokia Research Center, IBM T. J. Watson Research Center, HP Labs Japan, Hong Kong University of Science and Technology (HKUST), and Mitsubishi Electric Research Labs (MERL). His work has resulted in 10 U.S. patents granted with several pending. His research interests include signal, image, and video processing.
\end{IEEEbiography}
\end{document}